\documentclass[journal]{IEEEtai}

\usepackage[colorlinks,urlcolor=blue,linkcolor=blue,citecolor=blue]{hyperref}

\usepackage{color,array}

\usepackage{graphicx}

\usepackage{times}
\usepackage{soul}
\usepackage{url}
\usepackage{hyperref}
\usepackage[utf8]{inputenc}
\usepackage[small]{caption}
\usepackage{graphicx}
\usepackage{epstopdf}
\usepackage{amsmath}
\let\labelindent\relax
\usepackage{amsthm,bm,enumitem}
\usepackage{amsfonts}
\usepackage{booktabs}
\usepackage{multirow}
\usepackage{algorithmic}
\usepackage[ruled,vlined,linesnumbered]{algorithm2e}
\usepackage{bm}
\usepackage{hyperref}
\usepackage{textcomp}
\usepackage{xcolor}
\usepackage{subfigure}
\usepackage[switch]{lineno}
\theoremstyle{definition}

\theoremstyle{remark}
\newtheorem*{remark}{Remark}
\usepackage{setspace}
\usepackage{geometry}
\newgeometry{left=0.75in,right=0.75in,top=0.75in,bottom=1in}

\newtheorem{theorem}{Theorem}

\newcommand{\tabincell}[2]{\begin{tabular}{@{}#1@{}}#2\end{tabular}}

\begin{document}

\title{Vertical Federated Principal Component Analysis and Its Kernel Extension on Feature-wise Distributed Data}

\author{Yiu-ming Cheung, \IEEEmembership{Fellow, IEEE}, Juyong Jiang, Feng Yu, and  Jian Lou
% \thanks{This paragraph of the first footnote will contain the date on which you submitted your paper for review. It will also contain support information, including sponsor and financial support acknowledgment. For example, ``This work was supported in part by the U.S. Department of Commerce under Grant BS123456.'' }
\thanks{Yiu-ming Cheung, Juyong Jiang, and Feng Yu are with the Department of Computer Science, Hong Kong Baptist University, Hong Kong SAR, China (e-mail:ymc@comp.hkbu.edu.hk, csjyjiang@comp.hkbu.edu.hk, fengyu.sophia@gmail.com).}
\thanks{Jian Lou is with Xidian University, China (e-mail: jlou@xidian.edu.cn).}
\thanks{Yiu-ming Cheung is the Corresponding Author.}
% \thanks{This paragraph will include the Associate Editor who handled your paper.}
}

\markboth{}
{Y.M. Cheung \MakeLowercase{\textit{et al.}}}

\maketitle

\begin{abstract}
Despite enormous research interest and rapid application of federated learning (FL) to various areas, existing studies mostly focus on supervised federated learning under the horizontally partitioned local dataset setting. This paper will study the unsupervised FL under the vertically partitioned dataset setting. Accordingly, we propose the federated principal component analysis for vertically partitioned dataset (VFedPCA) method, which reduces the dimensionality across the joint datasets over all the clients and extracts the principal component feature information for downstream data analysis. We further take advantage of the nonlinear dimensionality reduction and propose the vertical federated advanced kernel principal component analysis (VFedAKPCA) method, which can effectively and collaboratively model the nonlinear nature existing in many real datasets. In addition, we study two communication topologies. The first is a server-client topology where a semi-trusted server coordinates the federated training, while the second is the fully-decentralized topology which further eliminates the requirement of the server by allowing clients themselves to communicate with their neighbors. Extensive experiments conducted on five types of real-world datasets corroborate the efficacy of VFedPCA and VFedAKPCA under the vertically partitioned FL setting. Code is available at \href{https://github.com/juyongjiang/VFedPCA-VFedAKPCA}{\textcolor{blue}{https://github.com/juyongjiang/VFedAKPCA}}.
\end{abstract}


\section{Introduction}

\IEEEPARstart{F}{ederated} learning (FL) \cite{ref1} has been receiving increasing attention in the literature, which enables collaborative machine learning in a communication-efficient \cite{shah2021model,xu2020ternary} and privacy-preserving way\cite{DBLP:journals/tnn/SattlerMS21}. FL provides a general-purpose collaborative learning framework, which usually consists of a coordinating central server and multiple participating clients with their local datasets, e.g., organizations (cross-silo setting) or devices (cross-device setting). More recent FL methods also propose fully-decentralized learning, where clients directly communicate with their neighbors. It eliminates the need of the coordinating server which can sometimes expose security and privacy risks to the collaborative training \cite{ref2}. During the FL training, the raw local datasets of all clients are kept locally and restricted to be exchanged or transferred over network for privacy-preserving purpose. Instead, it suffices to communicate only intermediate training variables (e.g., local gradients or local model parameters). Due to its generality and superiority in the privacy-preserving aspect, FL paves its way to a growing number of application areas.

Nevertheless, most existing FL methods focus almost exclusively on the supervised learning problems under the horizontally distributed local dataset setting. As far as we know, unsupervised FL learning under the settings of vertically distributed datasets has yet to be well explored. In fact, the vertically partitioned local dataset setting is common in many practical applications \cite{gu2021privacy}. Under this setting, different clients hold a local partition of features, instead of samples as in the horizontally partitioned setting. It naturally arises when the features/attributes describing the same group of entities are collected and stored among multiple clients. For example, the financial features of a person can split among multiple financial companies s/he has dealt with, e.g., banks, credit card companies, stock market. Similar to the horizontal FL setting, it is important to collaboratively train the model based on all clients' data partition to maximize the global model performance. For example, when a bank refers to a machine learning (ML) model for deciding whether to grant a loan application of a customer, it is ideal for the model training to take account into all the financial records of the customer, not only the transactions history held by this bank. It is apparent from the example that raw local datasets should not be exchanged because the vertically partitioned datasets can contain sensitive information, i.e., the financial status of the customer. Furthermore, unsupervised learning is practically appealing because it need not the labels for model training. The labels can be expensive and time-consuming to mark, and even require domain expertise \cite{wang2021unsupervised}. The label can also contain sensitive information, which will incur privacy leakage if not handled properly, especially under the vertically partitioned setting where the labels need to be shared among all clients \cite{sattler2021fedaux}.

In this paper, we focus on multiple clients holding the vertically partitioned datasets under the FL framework. For linear structure datasets, principal component analysis (PCA) has been widely used in high-dimensional data analysis by extracting principal components, which are uncorrelated and ordered, with the leading ones retaining most of the data variations \cite{ref5}. To this end, we propose a vertically partitioned federated PCA method, abbreviated VFedPCA, which features computational efficiency, communication efficiency and privacy-preserving. However, traditional PCA is limited to learning linear structures of data and it is impossible to determine dimensionality reduction when the data possesses nonlinear space structures. For nonlinear structure datasets, kernel principal component analysis (KPCA) is a very effective and popular technique to perform nonlinear dimensionality reduction \cite{scholkopf1998nonlinear}. But KPCA requires mapping data to high-dimensional sample space to obtain the final sample representations focusing on the feature-wise data, which is limited by the number of variables used, and the computational complexity will increase exponentially with the number of input variables. Therefore, we further propose the vertically partitioned federated KPCA method, namely VFedAKPCA. It first maps the data to the high-dimensional feature space and extracts the principal component directions, and then gets the final projection of the input data among those directions. In both VFedPCA and VFedAKPCA, clients keep their datasets local and only require the communication of model parameters. Such model exchange suffices to occur periodically to reduce communication overhead. Within each communication round, clients run the computationally efficient local power iteration \cite{ref22}, and the warm-start power iteration method can further improve performance. Furthermore, we consider the relationship between each independent subset and the full set based on the weight ratio of eigenvalue. Subsequently, we supplement the weight scaling method to further improve the accuracy and performance of the algorithm. Correspondingly, we test and verify the nonlinear feature extraction ability of the VFedAKPCA which has the same computational complexity as VFedPCA. In addition, we consider both of two settings: (1) a centralized server coordinates the learning, and (2) the fully decentralized setting which eliminates the need of the server.

\color{black} This paper is an extended version of our conference version, which contains the preliminary study only on the vertical federated PCA algorithm. In particular, the conference version mainly focuses on the one-shot communication performance of the proposed algorithm and the experiment results are light. This extended journal version is more comprehensive in terms of both algorithm design and experiments.\color{black} In summary, the main contributions of this paper are below:
\begin{enumerate}
    \item We first attempt to study PCA and KPCA under the vertically-partitioned FL setting.
    \item We propose the VFedPCA and VFedAKPCA algorithm for linear and nonlinear data situation, respectively, featuring computational efficiency, communication efficiency, and privacy-preserving. It comes with two variants: one requires a central server for coordination and the other performs fully decentralized learning where  clients communicate directly with their neighbors.
    \item Extensive experiments on five types of real datasets corroborate the favorable features of VFedPCA and VFedAKPCA under the vertically-partitioned FL setting, while maintaining accuracy similar to unseparated counterpart dataset.
\end{enumerate}

\section{Notations and Background}\label{2}
\subsection{Notation}\label{2.1}
We use boldfaced lower-case letters, e.g., $\mathbf{x}$, to denote vectors and boldfaced upper-case letters, e.g., $\mathbf{X}$, to denote matrix. The superscript is associated with the number of iterations, e.g., $\mathbf{X}^{t}$ denotes the decision variable at iteration $t$, while the subscript is associated with indices of different parties, e.g., the data matrix $\mathbf{X}_{i}\in \mathbb{R}^ {\mathnormal{n} \times \mathnormal{p}_i}$ denotes the i-th party that has $\mathnormal{p}_{i}$ variables (i.e., features) and $n$ samples (i.e., users). We use $\mathbf{X}^{*}$ to denote a dual space. Let $\mathbf{\|x}\|$ denote the standard Euclidean norm for the vector $\mathbf{x}$.
% For image dataset, the data matrix \(\mathbf{X}_{i} \in \mathfrak{R}^{n\times m\times m} \) is the i-th party's input dimension, where $m$ denotes the pixel size.
The summary of the used symbols in this paper is shown in Table \ref{t1}.
\setlength{\tabcolsep}{4pt}
\begin{table}
\begin{center}
\caption{Summary of Frequently Used Notation}
\label{t1}
\begin{tabular}{lc c c c|}
\hline\noalign{\smallskip}
Notation & Description\\
\noalign{\smallskip}
\hline
\noalign{\smallskip}
 $\mathnormal{n}$ & number of samples\\
 $\mathnormal{m}$ & number of features\\
 $\mathnormal{p}$ & partitions\\
 $l$ & local iterations\\
 $t$ & federated communications\\
 ${f}_{i}$ & the features of the i-th party\\
 ${s}_{i}$ & the samples of the i-th party\\
 $\alpha_{i}$ & the eigenvalues of the i-th party\\
 $\mathbf{a}_{i}$ & the eigenvectors of the i-th party\\
 $\omega_{i}$ & the weight of the i-th party\\
 $\mathbf{U}$ & the federated eigenvector\\
 $\mathbf{U}_G$ & the global eigenvector\\
\noalign{\smallskip}
\hline
\end{tabular}
\end{center}
\end{table}
\subsection{Federated Learning}\label{2.3}
FL is data-private collaborative learning, where multiple clients jointly train a global ML model with the help of a central coordination server. The goal of FL is achieved through the aggregation of intermediate learning variables, while the raw data of each client is stored locally without being exchanged or transferred. According to different data partition structures, the distribution can be generally categorized into two regimes, namely horizontally- and vertically-partitioned data \cite{fan2019distributed}. Currently, there is relatively less attention paid to the vertically partitioned data. Most of the existing cross-silo FL work is based on the supervised datasets, including trees \cite{ref16}, linear and logistic regression \cite{ref17,ref19}, and neural networks \cite{ref20}. These supervised FL works rely on the labels, which are expensive to acquire and require domain expertise. For example, diabetes patients may wish to contribute to the FL with their everyday health monitor data like glucose level and blood press. Since patients lack advanced medical expertise, their local data are often unlabeled without a medical expert's evaluation.  As far as we know, the current work on the unlabelled data and vertical learning has rarely explored under the FL context. The only existing related papers focus on the semi-supervised FL \cite{jeong2020federated} and weakly-supervised FL \cite{Jin2020TowardsUU}, respectively.

\subsection{PCA and Power Iteration}\label{2.2}
In this subsection, we describe the basic PCA setup and present the centralized Power Iteration \cite{ref21}. Let  $\mathbf{X}\in \mathbb{R}^{\mathnormal{n}\times\mathnormal{m}}$ be the global data matrix constituted as if all local partitions were centralized together. The goal of PCA is to find the top eigenvectors of the covariance matrix $\mathbf{A} = \frac{1}{n} \mathbf{X}^T\mathbf{X}\in\mathbb{R}^{\mathnormal{m}\times\mathnormal{m}}$, which is a symmetric positive semidefinite (PSD) matrix. We assume that $\mathbf{A}$ has eigenvalues $1\geq\alpha_{1}\geq\alpha_{2}\cdots\geq\alpha_{M}\geq0$ with corresponding normalized eigenvectors $\mathbf{a}_{1},\mathbf{a}_{2},\cdots,\mathbf{a}_{M}$, then $\mathbf{A}\mathbf{a}_m=\alpha_m\mathbf{a}_m$, for any $m\in\{1,\dots,M\}$.

\noindent \textbf{Power Iteration:} The power method \cite{ref21} is an efficient approach to iteratively estimating the eigenvalues and eigenvectors. The intuition of the method is by observing that $\mathbf{A}^{k}\mathbf{a}_m=(\alpha_m)^{k}\mathbf{a}_m$ for all $\mathnormal{k}\in\mathbb{N}$, which is based on the above relation $\mathbf{A}\mathbf{a}_m=\alpha_m\mathbf{a}_m$. In detail, let $\mathbf{a}^{(0)}$ be an initial vector. The power method \cite{ref21} estimates the top eigenvector by repeatedly applying the update step below:
\begin{equation}
\bm{\omega}=\mathbf{A}\mathbf{a}^{(k-1)},
\mathbf{a}^{(k)}=\frac{\bm{\omega}}{\|\bm{\omega}\|}, \label{eq1}
\end{equation}
where at each iteration, $\mathbf{a}^{(k)}$ will get closer to the top principal eigenvector $\mathbf{a}_{1}$. When $k \rightarrow\infty$,  $\mathbf{a}^{(k)}$ will converge to $\mathbf{a}_{1}$.

Power method has been extended to the FL setting for the horizontally partitioned data. Recently, \cite{ref22} proposes a federated SVD algorithm based on the local power iteration, which incorporates the periodic communication for communication-efficiency. Their so-called LocalPower method can save communication by $\mathbf{\mathnormal{l}}$ times to achieve $\epsilon$-accuracy. To guarantee the convergence, they assume the data matrix $\bm{X}$ is partitioned randomly and the local correlation matrix $\mathbf{A}_{i} = \frac{1}{\mathnormal{s}_{i}} \mathbf{X}_{i}^T\mathbf{X}_{i}$ is a good approximation to the global correlation matrix $\mathbf{A} = \frac{1}{n} \mathbf{X}^T\mathbf{X}$, which is equivalent to $\|\mathbf{A}-\mathbf{A}_{i}\|\leq\eta\|\mathbf{A}\|$, where $\mathbf{A}_{i}\in\mathbb{R}^{\mathnormal{m}\times\mathnormal{m}}$ is the $i$-th partition, $\eta$ that is as small as  $\epsilon$ is to bound the difference between $\mathbf{A}_{i}$ and $\mathbf{A}$. In addition, paper \cite{guo2021privacy} proposes the FedPower which further introduces more rigorous privacy protection under the notion of differential privacy.
As mentioned, both methods consider the horizontally partitioned data, leaving the federated PCA for vertically partitioned data untackled. \color{black} In the horizontal setting, the subproblems on the clients are separable because the data partitions on clients are identically independent distributed (i.i.d.) samples, which is less challenging since we do not need to consider the correlations among clients. However, in the vertical setting, the subproblems on the clients are not separable because the data partitions on clients are different feature subsets of the same groups of samples, which are correlated and the iid assumption can no longer be made. As a result, our targeted vertical PCA/KPCA Federated Learning is more challenging than the horizontal setting. \color{black} According to \cite{wu2018review}, the horizontally partitioned data is more challenging due to the tighter inter-correlation among features than those among samples, which motivates us to formally study the vertically partitioned federated PCA in this paper.
% However, our setting allows all parties to independently calculate the principal component vector based on the basic PCA method, and derive the federated principal component vector in the global center according to the weights of the parties, which can reduce the complexity of communication calculations and ensure the privacy of participants.

\begin{figure*}[t]
\centering
\includegraphics[width=0.9\linewidth]{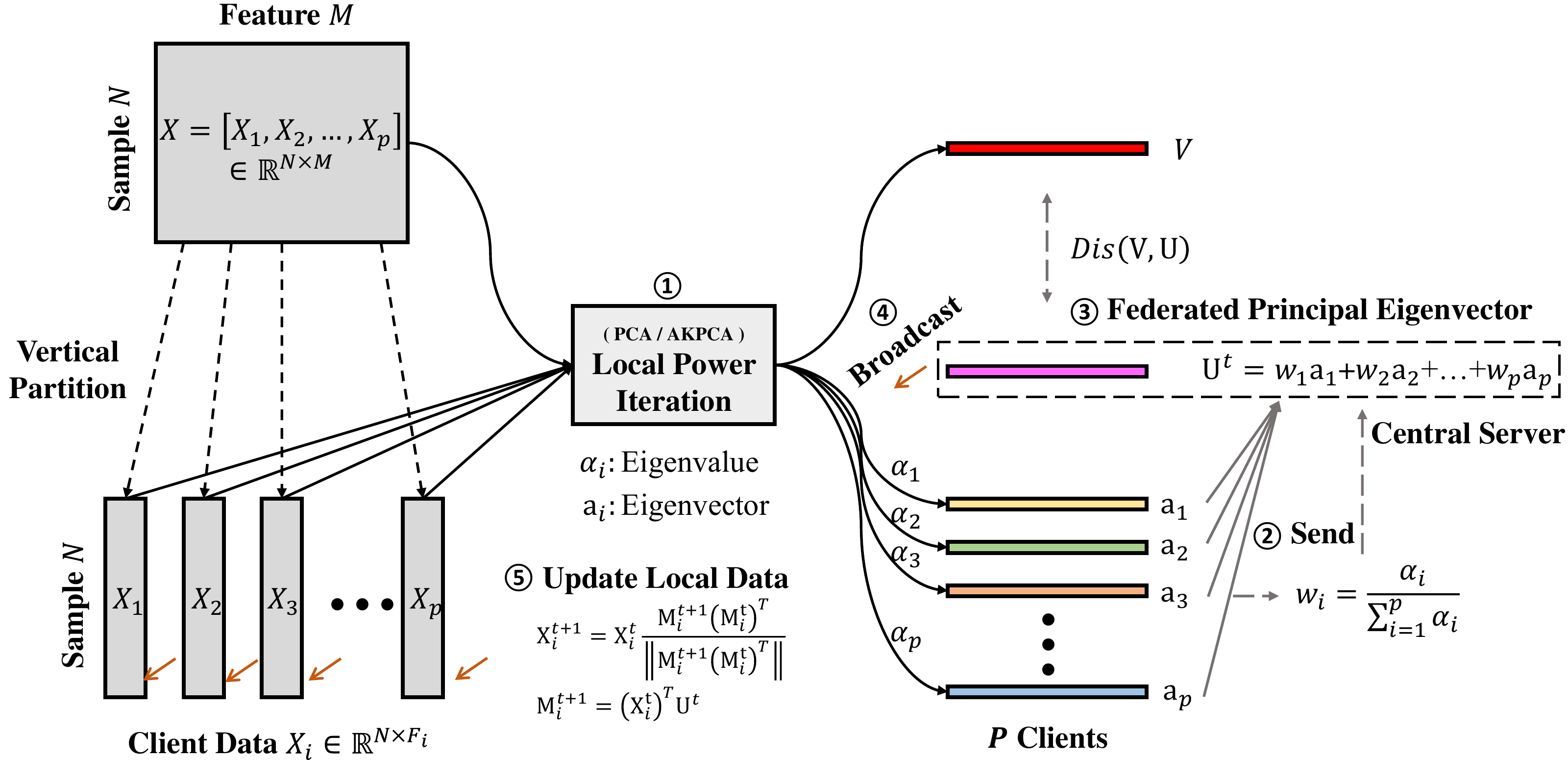}
\caption{The Architecture of Vertical Federated Principal Component Analysis.}
\label{fig:vfedakpca}
\end{figure*}

\begin{figure} \centering
\subfigure[] {\includegraphics[width=1\columnwidth]{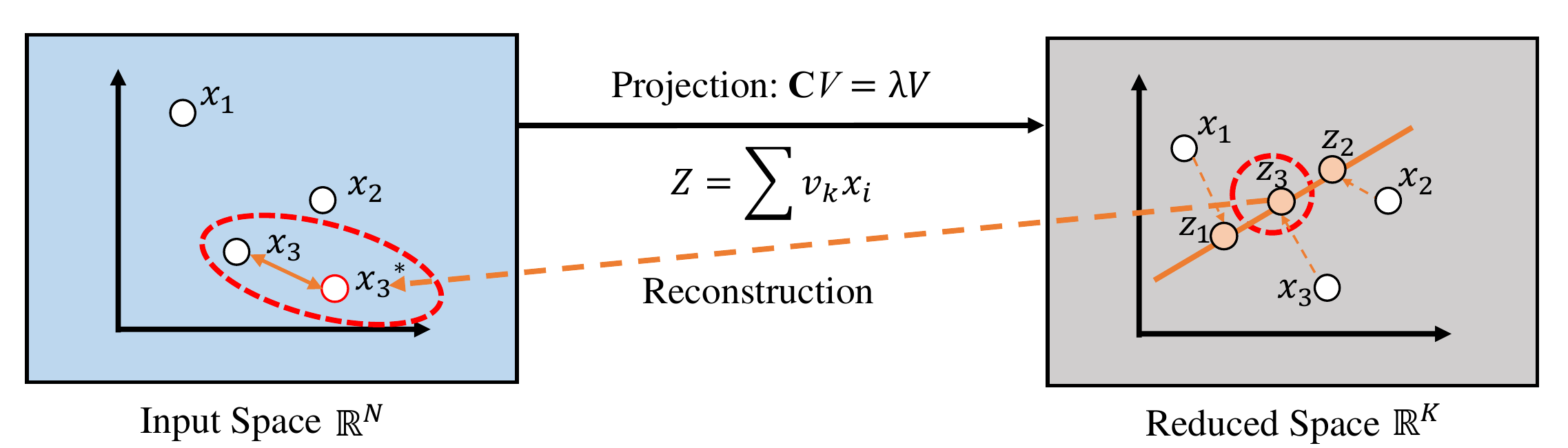}}
\subfigure[] {\includegraphics[width=1\columnwidth]{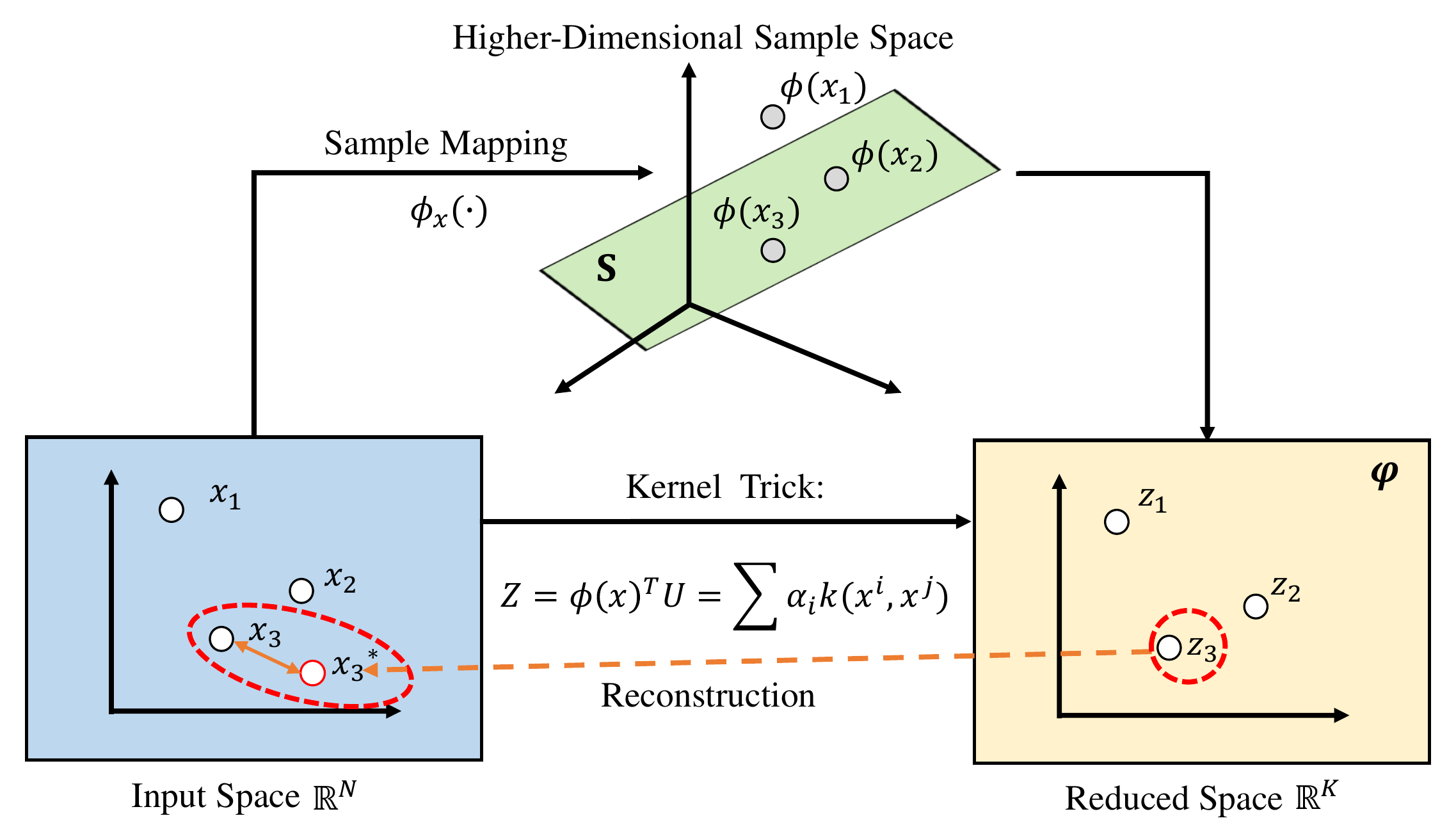}}
\subfigure[] {\includegraphics[width=1\columnwidth]{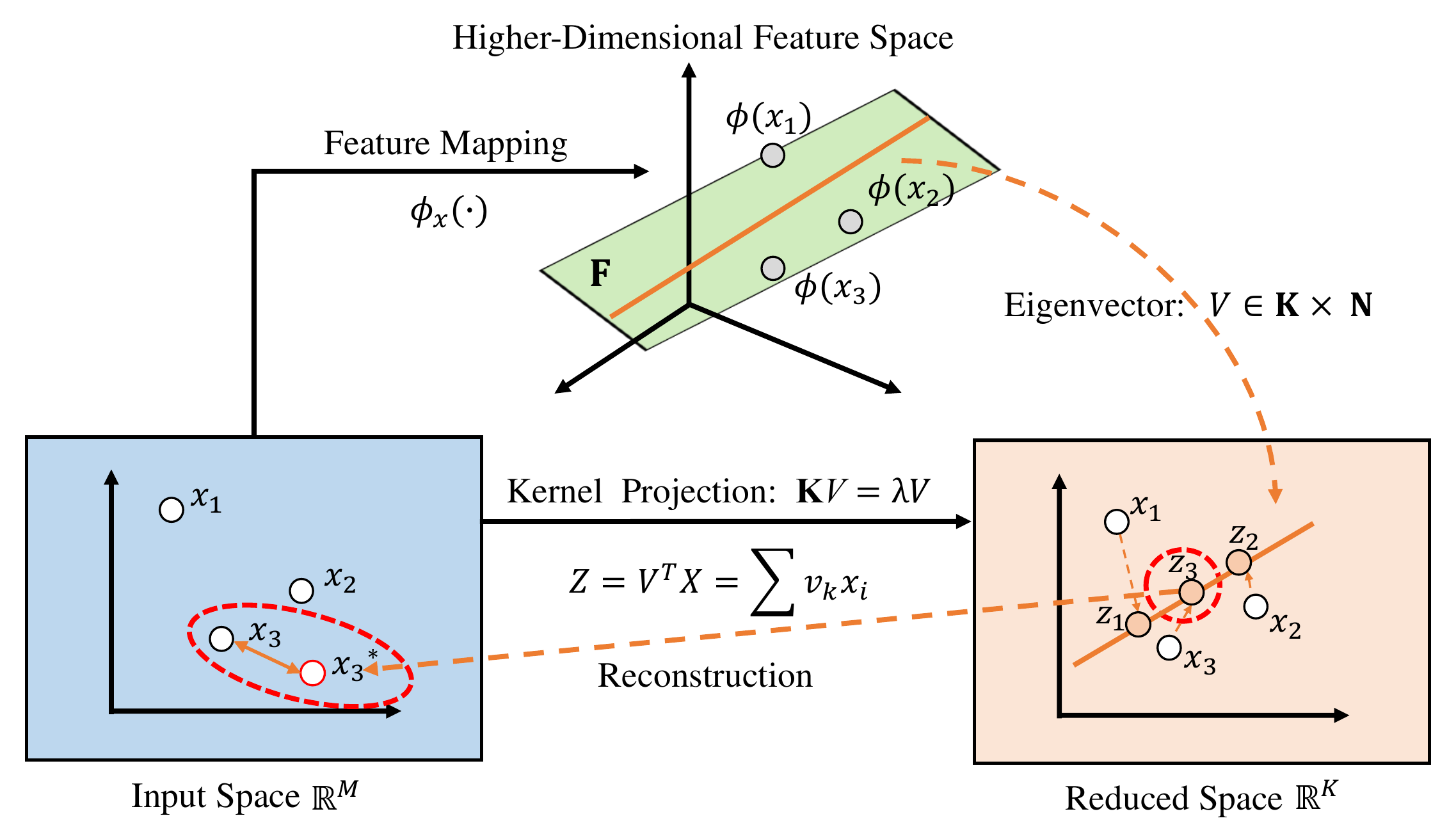}}
\caption{The illustrations of dimensionality reduction using (a) PCA (b) KPCA  in sample space (c) AKPCA from feature space to reduced space.}
\label{fig1}
\end{figure}

\subsection{Kernel PCA}
The kernel method has been widely recognized as an effective approach to processing nonlinear data through kernel mapping. In brief, the original data are first mapped to a higher-dimensional feature space by a kernel mapping function $\Phi: \mathbb{R}^{m} \rightarrow \mathbf{F}$, and then the corresponding linear operation is performed in this mapped feature space. By nonlinear mapping, linearly inseparable samples in the original space become linearly separable (or linearly separable with higher probability) in the higher-dimensional feature space. The linear transformation method can then be used to implement dimensionality reduction in the feature space, which greatly boosts the ability of linear transformation methods to process nonlinear data. The key principle behind this, also known as the kernel trick, is to exploit the fact that a great number of data processing techniques do not depend explicitly on the data itself, but rather on a similarity measure between them, i.e., an inner product \cite{honeine2011preimage}. Let $\{\Phi({x}_{1}), \Phi({x}_{2}), \dots, \Phi({x}_{n})\}$ denote the  input vectors $\{\mathbf{x}_{1},\mathbf{x}_{2},\dots\mathbf{x}_{n}\}$ after being mapped to the feature space, then kernel PCA (KPCA) is introduced to compute projections onto the eigenvectors $\mathbf{U}=\sum_{i=1}^{n} \mathbf{u}_{i}\Phi({x}_{i})$. We define an $\mathnormal{n}\times \mathnormal{n}$ kernel matrix $\mathbf{K}_{ij}=(\Phi({x}_{i}),\Phi({x}_{j}))$, then $\mathbf{u}_{i}=\frac{1}{\mathnormal{n}\lambda_{i}} \Phi({x}_{i})^{T} \mathbf{U}$, where $\lambda$ is an eigenvalue of $\mathbf{K}$.
% and \(\alpha\) is the associate eigenvector.
The resulting kernel principal components transformation is $\mathbf{\hat{X}}=\Phi({x}_{i})^{T} \mathbf{U}= \sum_{i=1}^{n}\mathbf{u}_{i}\mathbf{k}(\mathbf{x}_{i},\mathbf{x}_{j})$ may be called its nonlinear principal components corresponding to \(\Phi\), and the reduced dimension is $k\times n$.

KPCA could be disadvantageous if we need to process a very large number of observations because this results in a large matrix \(\mathbf{K}\). Compared to principal curves \cite{hastie1989principal},  which iteratively estimates a curve (or surface) capturing the structure of the data, KPCA is harder to interpret in the input space \cite{scholkopf1998nonlinear}.

\section{Vertical Federated PCA and Kernel PCA}\label{3}
%In this section, we first formulate the federated PCA and federated kernel PCA under the vertically partitioned data setting. Then, we propose the VFedPCA and VFedAKPCA algorithms to collaboratively extract principal components with the participation of all clients, while ensuring privacy by avoiding sharing of raw local datasets. Our method devises the local power iteration for efficient local computation, along with periodic communication for better communication efficiency. The overall framework of the VFedPCA(/VFedAKPCA) model is shown in Figure~\ref{fig:vfedakpca}.
\color{black}
In this section, we first formulate the federated PCA and federated kernel PCA under the vertically partitioned data setting. Then, we propose the VFedPCA and VFedAKPCA algorithms. Under the circumstance that each client does not know the eigenvector $\mathbf{U}_G$ of the mutual data (i.e., the overall dataset), the federated eigenvector $\mathbf{U}$ obtained by each client is finally converged to the global eigenvector $\mathbf{U}_G$ obtained from the overall data through federated learning. In addition, for the unsupervised learning approach, the key
challenge is to use the multi-shot federated interaction method to make the distance error between the federated eigenvector $\mathbf{U}$ and the global eigenvector $\mathbf{U}_G$ decrease continuously and converge gradually, that is mean, in the next federated interaction, each clients' $i$ can provide a better eigenvector $\mathbf{a}_i$. Our method devises the local power iteration for efficient local computation, along with periodic communication for better communication efficiency. The overall framework of the VFedPCA and VFedAKPCA model is shown in Figure~\ref{fig:vfedakpca}.
\color{black}
In particular, we consider two types of communication topologies. The first is the server-clients topology, which follows the most existing FL methods by introducing a semi-trusted server to coordinate the training. The second is the fully-decentralized topology, where the clients communicate in a peer-to-peer manner with their neighbors, which eliminates the need of the server that itself can be malicious and sometimes considered unpractical provided that such a semi-trusted server exist.

\subsection{Problem Formulation}
To begin with, we describe the formulation for the vertically federated PCA problem. Let $\mathbf{X}\in \mathbb{R}^{n\times m}$ be the data matrix, which have $n$ samples and $m$ features, we partition $\mathbf{X}$ into $p$ clients as $\mathbf{X}=[\mathbf{X}_{1},\mathbf{X}_{2},\cdots\mathbf{X}_{p}]^{T}$, where $\mathbf{X}_{i}\in \mathbb{R}^{n\times {f}_{i}}$ contains  $\mathnormal{f}_{i}$ features of $\mathbf{X}_{i}$. Let
$\mathbf{S} = \frac{1}{m} \mathbf{X}\mathbf{X}^{T}\in\mathbb{R}^{n\times n} $. Let $\mathbf{Z}=\mathbf{X}^{T}\mathbf{U} \in \mathbb{R}^{\mathnormal{m}}$, which can be considered as the coordinate of the projection along the direction given by  $\mathbf{U} \in \mathbb{R}^{\mathnormal{m}}$. Note that $\mathbf{Var}(\mathbf{Z})=\mathbf{U}^{T}\mathbf{S}\mathbf{U}$, where $\mathbf{S}$ is the covariance matrix. Our purpose is to find the leading $k$-dimensional subspace such that the projection of the data onto the subspace has the largest variance. The problem could be formulated as follows:
\begin{equation}
% \vspace{-0.2cm}
% \hspace{-0.1cm}
\underset{\mathbf{U} \in \mathbb{R}^{n \times k}, \mathbf{U}^{T} \mathbf{U}=\mathbf{I}}{\max}\|\mathbf{U}^{T}\mathbf{S}\mathbf{U}\|= \alpha_{G}\mathbf{U}^{T}_G\mathbf{U}_{G}=\alpha_{G},\label{eq3}
\end{equation}
where $\alpha_{G}$ are the leading eigenvalues of  $\mathbf{S}$ and $\mathbf{U}_{G}$ are the eigenvectors corresponding to $\alpha_{G}$.
Our aim is to minimize the distance between the global eigenvector $\mathbf{U}_G$ that is computed as if all features were centralized together, and each client’s $\mathbf{U}_{i}$, as follows
\begin{equation}
% \vspace{-0.2cm}
\underset{\mathbf{U}_G,\mathbf{U}_{i}\in\mathbb{R}^{\textit{m}\times\textit{k}}}{\min} dist(\mathbf{U}_G, \mathbf{U}_{i}),\text{ for all } i=1,...,p. \label{eq4}
\end{equation}
We define the $L$2 Euclidean distance \cite{ref29} as the distance between $\mathbf{U}_G$ and $\mathbf{U}_i$ by:
\begin{equation}
% \vspace{-0.2cm}
dist(\mathbf{U}_G, \mathbf{U}_{i}) =\sqrt{\sum_{j=1}^{m} \sum _{h=1}^k\left((\mathbf{U}_{G})_{j,h}-\mathbf{(U_\textit{i}})_{j,h}\right)^{2}},\label{eq5}
\end{equation}
where $\mathbf{U}_G$ is the leading eigenvectors of the global covariance matrix $\mathbf{S}$, and $\mathbf{U}_{i}$ is the leading eigenvectors of each client's (say $i$-th) local covariance matrix.

\subsection{Local Power Method}\label{3.2}
Let us consider the data vertically partitioned among $\mathnormal{p}$ clients. In each client, we use power iteration algorithm (also known as the power method) to produce the greatest (in absolute value) eigenvalue of $\mathbf{A}_{i}=\frac{1}{\mathnormal{f}_{i}}\mathbf{x}_{i}^{T}\mathbf{x}_{i}$, and a nonzero vector $\mathbf{a}_{i}$, which is a corresponding eigenvector of $\alpha_{i}$, i.e. $\mathbf{A}_{i}\mathbf{a}_{i} =\alpha_{i}\mathbf{a}_{i}$.  For the local iterations of the power method $\mathnormal{l} = 1,2 \dots \mathnormal{L}$, each client will compute locally until convergence (or the desired accuracy) by the steps
\begin{equation}
% \vspace{-0.2cm}
\mathbf{a}_{i}^{l+1} = \frac{\mathbf{A}_{i}\mathbf{a}_{i}^{l}}{\|\mathbf{A}_{i}\mathbf{a}_{i}^{l}\|},\label{eq6}
\end{equation}\\
where the vector $\mathbf{a}_{i}^{l}$ is multiplied by the matrix $\mathbf{A}_{i}$ and normalized at every iteration. Throughout the paper, we use $l\in[L]$ to denote local iterations and reserve $t\in[T]$ to denote global rounds.

If $\mathbf{a}_{i}^{l}$ is an (approximate) eigenvector of $\mathbf{A}_{i}$, its corresponding eigenvalue can be obtained as
\begin{equation}
% \vspace{-0.2cm}
\alpha_{i}^{l} = \frac{\mathbf{A}_{i}(\mathbf{a}_{i}^{l})^{T}\mathbf{a}_{i}^{l}}{(\mathbf{a}_{i}^{l})^{T}\mathbf{a}_{i}^{l}},\label{eq7}
\end{equation}\\
where $\mathbf{a}_{i}^{l}$ and $\alpha_{i}^{l}$ represent the largest eigenvector and eigenvalue of the $i$-th client, respectively.

\subsection{KPCA and Advanced KPCA method}\label{sec:c}
Let $\{\Phi({x}_{1}), \Phi({x}_{2}), \cdots, \Phi({x}_{m})\}$ denote the transformed data matrix after the kernel mapping. KPCA as a nonlinear feature extractor has proven powerful, yet has a bottleneck in terms of its computational complexity, which is mainly caused by the potentially large size of the kernel matrix. When focusing on the vertically partitioned data, our aim is to extract the leading eigen space as if the transformed data matrix was stored in the centralized manner.
% When the eigenvalue calculation of the kernel covariance matrix \(\mathbf{K}\) occurs in the space \(\mathbf{S}\), it may lead to expensive calculations due to process a large number of variables, and even cause the curse of dimensionality \cite{koppen2000curse}.
We propose the vertically federated advanced kernel PCA method, where 1) the kernel function is used to capture the nonlinear relationship between the data in the high-dimensional space; 2) Then, the principal component direction is extracted; 3) Finally, the original data is projected in this direction to obtain the ultimate feature extraction results. Our method avoids performing costly computations in a high-dimensional feature space, while only involves eigenvalue calculation on the kernel matrix $\mathbf{K}$, which can maintain the computational complexity close to PCA while still capable to extract the nonlinear characteristics.

The kernel function is the inner product term and denoted as the scalar
\begin{equation}
\mathbf{K}(\mathbf{x},\mathbf{x}_{i})=\Phi({x})^{T}\Phi({x}_{i}).\label{eq20}
\end{equation}
By computing $\mathbf{K}(\mathbf{x}_{i},\mathbf{x}_{j})$ as the $ij$-th element of the  $n \times n$  matrix $\mathbf{K}$, we have
\begin{equation}
\mathbf{K}=\{\mathbf{K}(\mathbf{x}_{i},\mathbf{x}_{j})\}=\{\Phi({x}_{i})^{T}\Phi({x}_{j})\}_{i,j=1,2,\ldots,n}\label{eq21}
\end{equation}
% The covariance matrix \(\mathbf{K}\) in is
% \begin{equation}
% \mathbf{K}=\frac{1}{n}\sum_{j=1}^{n} \Phi({x}_{j})\Phi({x}_{j})^{T}\label{eq22}
% \end{equation}
It is now suffice to find eigenvalues $\lambda\geq0$ and eigenvectors $\mathbf{V}$ which satisfies
\begin{equation}
	\mathbf{K}\mathbf{V} = \lambda\mathbf{V}, \label{eq23}
\end{equation}
where the eigenvectors $\mathbf{V}=[\mathbf{v}_{1},\mathbf{v}_{2},\ldots,\mathbf{v}_{n}]$.
Then, we project the input features on eigenvectors, the final principal components features result is
\begin{equation}
\mathbf{Z} = \mathbf{V}^T \mathbf{X} = \sum_{i=1}^{n} \mathbf{v}_{k}^{T}\mathbf{x}_{i},
% \mathbf{\hat{X}}=\Phi({x}_{i})^{T} \mathbf{V}.
\label{eq24}
\end{equation}
where $\mathbf{Z}=[\mathbf{z}_{1},\mathbf{z}_{2},\ldots,\mathbf{z}_{n}] \in  \mathbb{R}^{\mathnormal{k}\times \mathnormal{m}}$.

The illustrations of dimensionality reduction by using the three methods (PCA, KPCA and AKPCA) are shown in Figure \ref{fig1}.

\subsection{Kernel Selection}
\begin{figure}[t]
\centering
\includegraphics[width=1\columnwidth]{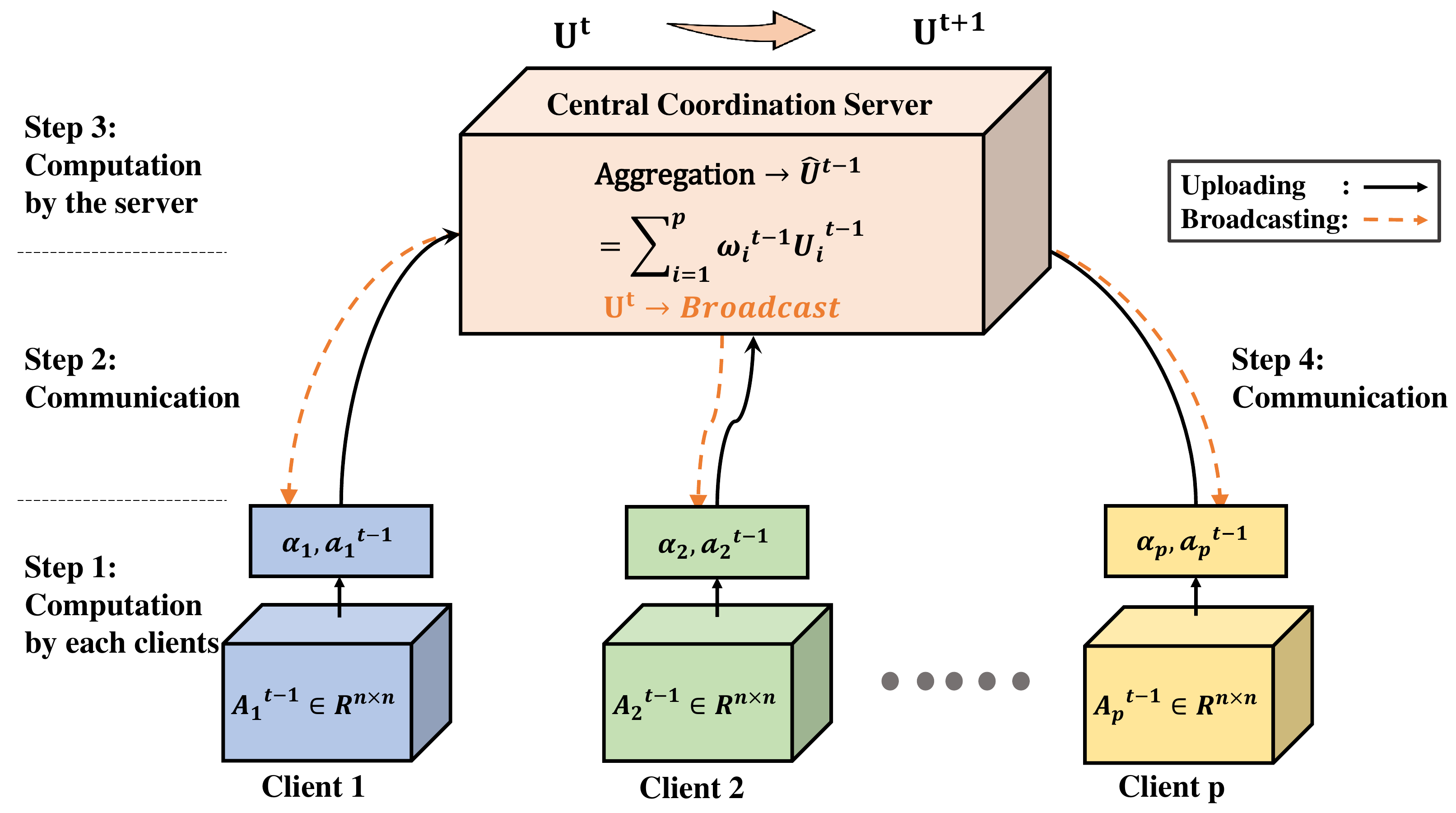}
\caption{The Framework of Sever-Clients Architecture.}
\label{fig:sc_arc}
\end{figure}

The powerful kernel methods eliminates the need to compute $\mathbf{\Phi}_{i}$ explicitly, where $\mathbf{\Phi}$ is a nonlinear function with the only  structural restriction of being a reproducing kernel Hilbert space (RKHS) \cite{aronszajn1950theory}. Rather, we can directly construct the kernel matrix from the input dataset $\{\mathbf{x}_{i}\}$ (Weinberger et al., 2004) \cite{weinberger2004learning}. Two commonly used kernels are exemplified in this paper, while other choices of kernels are also well-accommodated by our VFedAKPCA approach: (1) the radial basis function (RBF) kernels, which are functions of $\mathbf{x}_{i}-\mathbf{x}_{j}$, and (2) the sigmoid kernels, which depend on the dot product of the input arguments $\mathbf{x}_{i}$ and $\mathbf{x}_{j}$.

\(\bullet \) RBF (Radial Basis Function) kernel, sometimes referred to as the Gaussian kernel:
\begin{equation}
\mathbf{k}(\mathbf{x}_{i},\mathbf{x}_{j})=\mathbf{exp}(-\gamma \|{\mathbf{x}_{i}-\mathbf{x}_{j}}\|^2), \label{eq25}
\end{equation}
where $\gamma=\frac{1}{2\sigma^2}$.
The space constructed from Gaussian kernels is infinite-dimensional for input vectors, since every input $\mathbf{x}$ maps to a Gaussian function being linearly independent from the mapped space of a distinct $\mathbf{x}_{i}$.
% Where the input vectors in KPCA and AKPCA are \(\mathbb{R}^{n}\) and \(\mathbb{R}^{m}\), respectively.

\(\bullet \) Sigmoid kernel
\begin{equation}
\mathbf{k}(\mathbf{x}_{i},\mathbf{x}_{j})=\mathbf{tanh}(-\gamma\mathbf{x}_{i}{\mathbf{x}_{j}}^{T} + c), \label{eq26}
\end{equation}
where $c\geq0$ and $\gamma=\frac{1}{2\sigma^2}$. Sigmoid kernels are not positive definite and therefore do not induce a reproducing kernel Hilbert space, yet they have been successfully used in practice \cite{scholkopf1998nonlinear}.

\subsection{Federated Communication in Server-Clients Architecture}\label{3.3}
%Under the vertically partitioned data setting, the overall pipeline of the VFedPCA(/VFedAKPCA) model is shown in Figure~\ref{fig:vfedakpca}. Each clients' $\mathbf{X}_{i}$ calculates the local power iteration locally, then share the eigenvalue $\alpha_{i}$ and eigenvector $\mathbf{a}_{i}$ publicly. Our aim is to minimize the distance between the global eigenvector $\mathbf{U}_{G}$ and the federated eigenvector $\mathbf{U}$ continuously. In particular, the core steps of federated communication in server-clients architecture is summarized as follows:

\color{black} With the local kernelization as obtained by Algorithm \ref{algorithm3}, we are ready to derive our federated algorithms, which are separated into Server-Client architecture and Fully Decentralized architecture. \color{black}

\begin{algorithm}
\textbf{Input:} $\Phi(x)\in\mathbb{R}^{n}$.

\textbf{Output:}  $\mathbf{x}\in\mathbb{R}^{k}$ where $\mathnormal{k} \ll \mathnormal{n}$
\SetAlgoLined

Select a kernel and calculates the matrix $\mathbf{K}$:

$\mathbf{K}(\mathbf{x},\mathbf{x}_{i})=\Phi({x})^{T}\Phi({x}_{i})$

Find the eigenvector $\mathbf{V}$ of the kernel matrix $\mathbf{K}$:

$\mathbf{K}\mathbf{V} = \lambda\mathbf{V}$

Project the data among the first $k$ eigenvectors:

%\(\mathbf{\hat{X}}=\Phi({x}_{i})^{T} \mathbf{V}\)
$\mathbf{Z} = \mathbf{V}^T \mathbf{X} = \sum_{i=1}^{n} \mathbf{v}_{k}^{T}\mathbf{x}_{i}$

\textbf{return} final results.
\caption{Advanced KPCA Based on Vertically Partitioned Dataset Procedure}
\label{algorithm3}
\end{algorithm}

\textbf{\emph{Model integration:}}
Each client uploads the eigenvector $\mathbf{a}_{i}^{L}$ and the eigenvalue $\alpha_{i}^{L}$ to the central sever. First, the server computes the weight $\omega_{i}$ of each client $i$. Then, the server merges all clients' results and computes the federated eigenvector $\mathbf{u}^{t}$ as follows,
\begin{equation}
\mathbf{u}^{t} = \omega_{1}^{t}\mathbf{a}_{1}^{t} + \omega_{2}^{t}\mathbf{a}_{2}^{t} + \cdots \omega_{p}^{t}\mathbf{a}_{p}^{t} \label{eq9},
\omega_{i}^{t} = \frac{\alpha_{i}^{t}}{\sum_{i=1}^{p}\alpha_{i}^{t}},
\end{equation}

where $\omega_{i}^{t}$ is the $i$-th client's weight and $\mathbf{u}^{t}$ is the shared projection feature vector by merging all clients' $\mathbf{a}_{i}^{t}$.

\textbf{\emph{Parameters broadcasting and updating:}}
Central coordination server broadcast the federated principal eigenvector $\mathbf{u}^t$ to each clients' $i$. Then each clients' $i$ receive the $\mathbf{u}^t$ and update the local dataset $\mathbf{X}_{i}^{t}$ to reach the better local eigenvector as follows,
\begin{equation}
    %\mathbf{X}_{i}^{t} \leftarrow \mathbf{u}^{t}(\mathbf{u}^{t})^{T} \mathbf{X}_{i}^{t}, t\in[T]
\mathbf{X}_{i}^{t} \leftarrow \mathbf{X}_{i}^{t} \frac{\mathbf{M}_{i}^{t}(\mathbf{M}_{i}^{t})^{T}}{\|\mathbf{M}_{i}^{t}(\mathbf{M}_{i}^{t})^{T}\|},
\mathbf{M}_{i}^{t} \leftarrow (\mathbf{X}_i^{t})^{T}\mathbf{u}^{t}, t\in[T]
\end{equation}
where $t\in[T]$ is to denote the multi-shot rounds.
%It comes with two variants: (1) one requires a central server for coordination and (2) one performs fully decentralized learning where each clients' $i$ communicate directly with their neighbors.

\subsection{Overall Pipeline of the Sever-Clients Architecture}\label{3.5}

All clients share parameters with the help of the central coordination server. In our setting, this third-client coordinator can be trusted, which means that it will honestly conduct the designated functionality and will not attempt to breach the raw data of any clients. After adjusting the new parameters, this federated-parameter is returned to all clients. Then, each client calculates and updates the local datasets based on this new federated-parameter locally. Moreover, this communication-local computation cycle is iterative. The framework of this model is shown in Figure~\ref{fig:sc_arc}. The algorithm is summarized in Algorithm \ref{a1}. The step of this method is as follows:

\textbf{\emph{Local training:}} All clients involved perform local power method independently. Each client first calculates the covariance matrix $\mathbf{A}$. Subsequently the largest eigenvector $\mathbf{a}$ and eigenvalue $\alpha$ are calculated (see steps 4-10 of Algorithm \ref{a1});

\textbf{\emph{Model integration:}} The central server calculates the weight $\omega_{i}$ occupied by each client in the federated model by receiving the eigenvalue $\alpha_{i}$ of each client, and then combines the received eigenvector $\mathbf{a}_{i}, i \in \{1, \cdots p\}$, to derive a federated feature vector $\mathbf{u}^{t}$, each client update the local data, where the communication is cyclical (see steps 17-23 of Algorithm \ref{a1});

\textbf{\emph{Parameters broadcasting and updating:}} The central coordination server broadcasts the aggregated parameters $\mathbf{u}^{t}$ to the $\mathnormal{p}$ clients (see step 22 of Algorithm \ref{a1}). Each client updates local eigenvector with the new returned federated feature vector $\mathbf{u}^{t}$ and updates the local data (see step 11-15 of Algorithm \ref{a1}).

The advantage of this model is relatively efficient, although it relies on the help of the server.
%\subsection{Sever-Clients Architecture}\label{3.4}

\begin{algorithm}[t]
\SetAlgoLined
\textbf{Input:} Data $\mathbf{X}_{i}=\{\mathbf{x}_{1}, \mathbf{x}_{2}, \cdots, \mathbf{x}_{f_{i}}\} \in \mathbb{R}^{n\times f_{i}}$ belongs to client (node) $i$

\textbf{Output:} Federated principal eigenvector \(\mathbf{u} \)

\textbf{$\Rightarrow$ Run on the $i$-th node}

\textbf{Initial:} Let $\mathbf{a}_i^{(0)} \in  \mathbb{R}^ {n}$ randomly

\For{$l$ = 0 to $L-1$}{
\While{\textit{each client} $i \in \{1,2 \cdots p\}$}
{
    % $\mathbf{A}_{i}\leftarrow \mathbf{X}_{i}$,

    $\mathbf{a}_{i}^{l+1} = \frac{\mathbf{A}_{i}\mathbf{a}_{i}^{l}}{\|\mathbf{A}_{i}\mathbf{a}_{i}^{l}\|}$,
    $\alpha_{i}^{l+1} = \frac{\mathbf{A}_{i}(\mathbf{a}_{i}^{l})^{T}\mathbf{a}_{i}^{l}}{(\mathbf{a}_{i}^{l})^{T}\mathbf{a}_{i}^{l}}$
}
}
Send ($\alpha_{i}^{L}$, $\mathbf{a}_{i}^{L}$) to the central server.

Receive updated global aggregation from the server. Update client $i$'s local data:

\For{$i$ = 1 to $p$}{
 % $\mathbf{X}_{i}^{t} \leftarrow \mathbf{u}^{t}(\mathbf{u}^{t})^{T} \mathbf{X}_{i}^{t}$
  $\mathbf{M}_{i}^{t} \leftarrow (\mathbf{X}_i^{t})^{T}\mathbf{u}^{t}$

$\mathbf{X}_{i}^{t} \leftarrow \mathbf{X}_{i}^{t} \frac{\mathbf{M}_{i}^{t}(\mathbf{M}_{i}^{t})^{T}}{\|\mathbf{M}_{i}^{t}(\mathbf{M}_{i}^{t})^{T}\|}$
  }

\textbf{\(\Rightarrow\) Run on central coordination server}

Receive $\alpha_{i}^{L}$, $\mathbf{a}_{i}^{L}$ from $p$ clients

\For{$t$ = 0 to $T-1$}{
\For{$i$ = 1 to $p$}{
  $\mathbf{u}^{t+1}\leftarrow \text{Merge} (\omega_{i}^{t},  \mathbf{a}_{i}^{t})$,
  $\omega_{i}^{t+1} \leftarrow \alpha_{i}^{t}$
  }

Broadcast $\mathbf{u}^{t+1}$ to client $i$.

}
\caption{VFedPCA Learning with Central Coordination Server}
\label{a1}
\end{algorithm}

\subsection{Local Power Iteration with Warm Start}\label{3.5}
The general power iteration algorithm starts with a initialization vector $\mathbf{a}^{(0)}$, which may be an approximation to the dominant eigenvector or a random unit vector. Here, we initialize the algorithm from a “warm-start” vector $\mathbf{a}^{(0)}$ , which is based on the global aggregation of the previously communication round. It in practice helps the local clients to reach designated accuracy of the local power method, which will then reduce the running time and overall improve the performance of the local power iteration algorithm in Section \ref{3.2}. The algorithm is summarized in Algorithm \ref{a3}. The main step is as follows:

\textbf{\emph{Local training:}} Each node uses the federated feature vector $\mathbf{u}$ as the initialize value to perform local power method, then calculate the $k+1$ to $2k$ eigenvector $\mathbf{a}$ and eigenvalue $\alpha$ and send to the server (see steps 4-10 of Algorithm \ref{a3});

\begin{algorithm}[t]
\SetAlgoLined
\textbf{Input:} Data $\mathbf{X}_{i}=\{\mathbf{x}_{1}, \mathbf{x}_{2}, \cdots, \mathbf{x}_{f_{i}}\} \in \mathbb{R}^ {n\times f_{i}}$ belongs to client (node) $i$

\textbf{Output:} Client $i$ principal eigenvalue $\alpha_{i}^{l}$, and eigenvector $\mathbf{a}_{i}^{l}$

\textbf{\(\Rightarrow\) Run on the $i$-th node}

\textbf{Initial:} Let  $\mathbf{a}_i^{(0)} = \mathbf{u}$ with warm-start

\For{$l$ = 0 to $L-1$}{

\While{each client $i \in \{1,2 \cdots p\}$}{

$\mathbf{a}_{i}^{l+1} = \frac{\mathbf{A}_{i}\mathbf{a}_{i}^{l}}{\|\mathbf{A}_{i}\mathbf{a}_{i}^{l}\|}$, $\alpha_{i}^{l+1} = \frac{\mathbf{A}_{i}(\mathbf{a}_{i}^{l})^{T}\mathbf{a}_{i}^{l}}{(\mathbf{a}_{i}^{l})^{T}\mathbf{a}_{i}^{l}}$
}
}
Send $(\alpha_{i}^{L},\mathbf{a}_{i}^{L})$ to the central server.
\caption{Local Power Iteration with Warm Start}
\label{a3}
\end{algorithm}

\begin{algorithm}[t]
\SetAlgoLined
\textbf{Input:} Data $\mathbf{X}_{i}=\{\mathbf{x}_{1}, \mathbf{x}_{2}, \cdots, \mathbf{x}_{f_{i}}\} \in \mathbb{R}^ {n\times f_{i}}$ belongs to client (node) $i$

\textbf{Output:} Federated principal eigenvector $\mathbf{u}$

\textbf{\(\Rightarrow\) Run on central collaborative server}

\For{$t$ = 0 to $T-1$}{
Receive $\alpha_{i}^t,\mathbf{a}_{i}^t$ from $n$ clients

\For{$i$ = 1 to $p$}{

  $\omega_{i}^{t} \leftarrow \alpha_{i}^{t} $,
  $\eta_{i}^{t} = \frac{1}{p}\sum _{i=1}^{p} \omega_{i}^{t} $,

 $\mathbf{u}^{t+1}\leftarrow \text{Merge} (\eta_{i}^{t}, \omega_{i}^{t}, \mathbf{a}_{i}^{t})$}
Broadcast $\mathbf{u}^{t}$ to client $i$.

% Update client $i$'s local dataset:

% \For{$i$ = 1 to $p$}{
% %   $\mathbf{X}_{i}^{t} \leftarrow \mathbf{u}^{t}(\mathbf{u}^{t})^{T} \mathbf{X}_{i}^{t}$
%   $\mathbf{M}_{i}^{t} \leftarrow (\mathbf{X}_i^{t})^{T}\mathbf{u}^{t}$

% $\mathbf{X}_{i}^{t} \leftarrow \mathbf{X}_{i}^{t} \frac{\mathbf{M}_{i}^{t}(\mathbf{M}_{i}^{t})^{T}}{\|\mathbf{M}_{i}^{t}(\mathbf{M}_{i}^{t})^{T}\|}$
%  }
}

\caption{Weight Scaling Method}
\label{a4}
\end{algorithm}

% This method considers the use of an adaptive loop idea to allow communication between clients to reduce the error to a certain extent, especially for the case where the error between the general federated result \(\mathbf{u}\) and the global result \(\mathbf{u}_G\) is large.
\subsection{Weight Scaling Method}\label{3.6}
The federated result may be sometimes not beneficial to all clients.
% The weight factors used in preceding sections that determine the direction of federated eigenvector in the same sample space are based on the size of the eigenvalue of each client.
In general, the client with the larger eigenvalue has a greater influence on the federated result, which hints a natural intuition that following the direction of the clients with larger eigenvalues tends to reach the global consensus faster and costs less communication rounds. Inspired by this intuition, we introduce a weight scaling factor to further improve the federation communication. The improved weight scaled federated average is formulated by
\begin{equation}
\label{eq10}
\begin{split}
\mathbf{u}^{t} & = (1+\eta_{1}^t)\omega_{1}^{t}\mathbf{a}_{1}^{t} + \cdots + (1+\eta_{\lceil p/2\rceil}^t)\omega_{\lceil p/2\rceil}^{t}\mathbf{a}_{\lceil p/2\rceil}^{t} \\
& + (1-\eta_{\lceil p/2\rceil +1}^t)\omega_{\lceil p/2\rceil +1}^{t}\mathbf{a}_{\lceil p/2\rceil +1}^{t} \cdots (1-\eta_{p}^t)\omega_{p}^{t}\mathbf{a}_{p}^{t},
\end{split}
\end{equation}
where %$\sum _{i=1}^{\lceil p/2\rceil} \eta _{i}^t = \sum ^{p}_{i= \lceil p/2\rceil +1} \eta _{i}^t$.
$\eta_{i}^t = \frac{1}{p}\sum _{i=1}^{p} \omega_{i}^{t}$.
In brief, we gradually increase the impact of the first half of the clients with larger eigenvalues, while further decrease the impact of the clients of the second half with smaller eigenvalues.
The algorithm is summarized in Algorithm \ref{a4}. The main step is as follows:

\textbf{\emph{Model integration:}} The central server calculates the weight $\omega_{i}$ occupied by each client in the federated model based on the received eigenvalue $\alpha_{i}$ of each client, and then adds $\eta$ parameter to further adjust the weight scale of each client. Then, it combines the received eigenvector $\mathbf{a}_{i}, i \in \{1, \cdots p\}$, to derive a federated feature vector $\mathbf{u}^{t}$ (i.e., steps 4-11 of Algorithm \ref{a4});

This method is especially suitable for situations where some clients have a larger weight in the federated aggregation.

\begin{figure}[t]
\centering
\includegraphics[width=0.8\columnwidth]{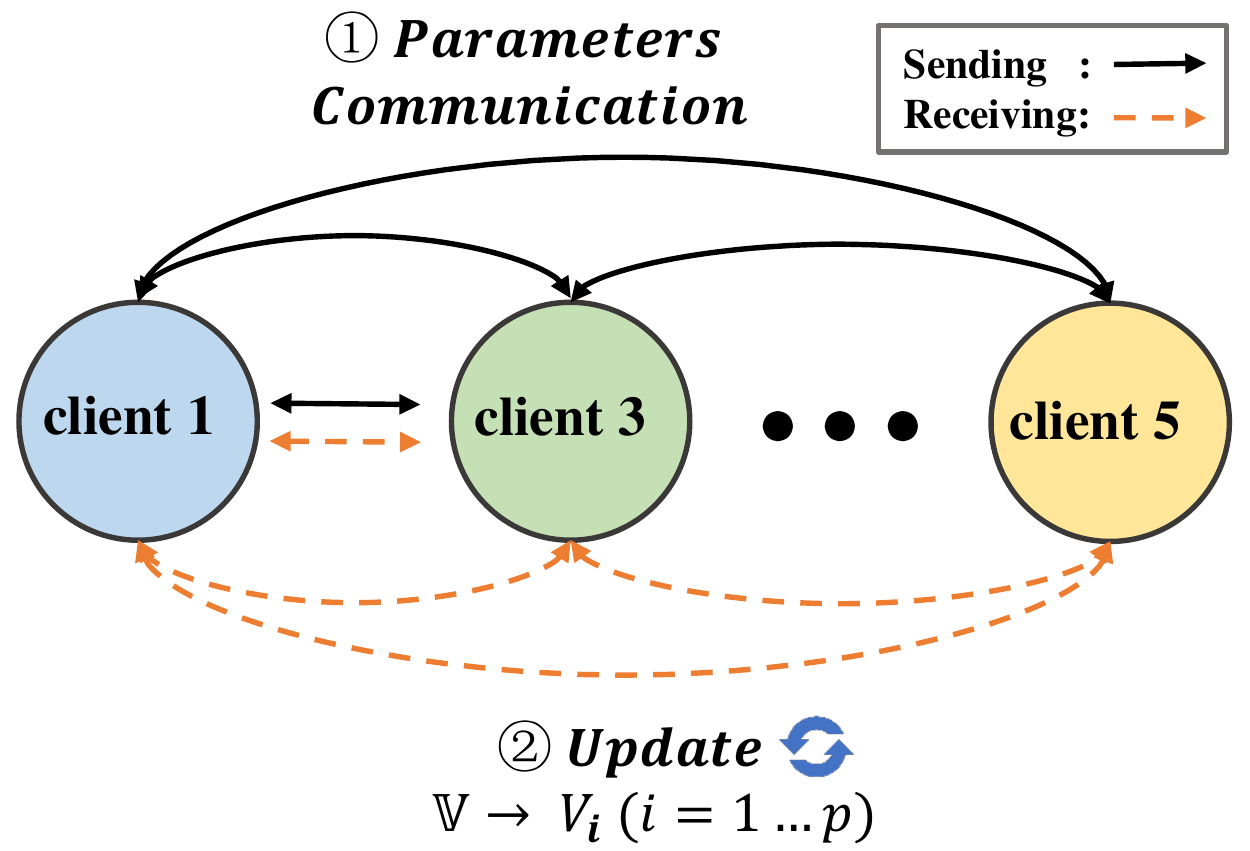}
\caption{The Framework of Fully Decentralized ({\it peer-to-peer}) VFPCA learning.}
\label{fig:example2}
\end{figure}

\subsection{Fully Decentralized Architecture}\label{3.6}

Next, we consider the fully decentralized setting, where all clients only share principal component parameters with their direct neighbors and no central server is required. The framework of this model is shown in Figure~\ref{fig:example2}. \color{black} Since both settings arise in different application scenarios, we do not intend to use one to take the place of the other, but to study both settings in order to be more comprehensive. As for the pros and cons, it is common to all Federated Learning with the different two settings. That is, the decentralized setting provides better privacy and does not rely on the trusted server assumption, while server-client architecture is relatively easier to coordinate but the server itself can bring about security and privacy issues. \color{black} The algorithm is summarized in Algorithm \ref{a2}. The main step of this method is as follows:

\textbf{\emph{Parameters communication and updating:}} Communication between clients connected to each other and share  intermediate results of the parameters $\alpha_{i}$ and $\mathbf{a}_{i}$. Each client obtains the extracted final data based on calculating the weights $\omega_{i}$'s occupied by all clients and updates locally (see steps 4-15 of Algorithm \ref{a2}).

This model circumvents the need for third-client agencies to help participants further reduce some external risks, especially for two-client scenarios. However, the principal components of all clients need to be calculated independently, and the computing efficiency will be greatly affected.

\begin{algorithm}[t]
\SetAlgoLined
\textbf{Input:} Data $\mathbf{X}_{i}=\{\mathbf{x}_{1}, \mathbf{x}_{2}, \cdots, \mathbf{x}_{f_{i}}\} \in \mathbb{R}^ {n\times f_{i}}$ belongs to client (node) $i$

\textbf{Output:} Client $i$ federated principal eigenvector $\mathbf{u}$

\textbf{\(\Rightarrow\) Run on the $i$-th node}

\# Parameters communication between connected clients

\For{$t$ = 1 to $T$}{
\For{$i$ = 1 to $p$}{
\For{$l$ = 1 to $L$}{

$\mathbf{a}_{i}^{l} = \frac{\mathbf{A}_{i}\mathbf{a}_{i}^{l}}{\|\mathbf{A}_{i}\mathbf{a}_{i}^{l}\|}$,
%\(\alpha_{i}^{l} = \frac{\mathbf{A}_{i}\mathbf{a}_{i}^{l}\mathbf{a}_{i}^{l*}}{\mathbf{a}_{i}^{l}\mathbf{a}_{i}^{l*}}\)
$\alpha_{i}^{l} = \frac{\mathbf{A}_{i}(\mathbf{a}_{i}^{l})^{T}\mathbf{a}_{i}^{l}}{(\mathbf{a}_{i}^{l})^{T}\mathbf{a}_{i}^{l}}$
}
  $\mathbf{u}^{t}\leftarrow \text{Merge} (\omega_{i}^{t}, \mathbf{a}_{i}^{t})$,
  $\omega_{i}^{t} \leftarrow \alpha_{i}^{t}$
}
Update client $i$'s local dataset: %$\mathbf{X}_{i}^{t} \leftarrow \mathbf{u}^{t}(\mathbf{u}^{t})^{T} \mathbf{X}_{i}^{t}$

$\mathbf{M}_{i}^{t} \leftarrow (\mathbf{X}_i^{t})^{T}\mathbf{u}^{t}$

$\mathbf{X}_{i}^{t} \leftarrow \mathbf{X}_{i}^{t} \frac{\mathbf{M}_{i}^{t}(\mathbf{M}_{i}^{t})^{T}}{\|\mathbf{M}_{i}^{t}(\mathbf{M}_{i}^{t})^{T}\|}$
}
\caption{Fully Decentralized VFedPCA learning}
\label{a2}
\end{algorithm}

\subsection{Privacy and Complexity Analysis}\label{3.8}
\subsubsection{Privacy Analysis}
\color{black} For the privacy-preserving capability, our method provides the standard privacy protection of the federated learning setting, i.e., clients' raw data remain local and only model/model updates are communicated. In addition, we also studies the fully decentralized communication architecture to further enhance privacy. That is, it eliminates the need of the central server, thus enhancing the privacy in case the server exhibiting malicious behaviors. \color{black}

\subsubsection{Complexity Analysis}
In each client, let $L$ be the total number of local iterations, which only depends on the eigen-gap $\Delta$ after full passes over the data between the top two eigenvalues of $\mathbf{A}_{i}$.  In Algorithm 1, let $T$ be the total number of federated aggregations executed by the server.

\begin{theorem} \textbf{(Local Iteration Complexity)}
Assume that the initialization $\bm{a}_i^0$ is within $O(\frac{\alpha _i(1)-\alpha _i(2)}{\sqrt{\alpha _i(1)}})$ distance from $\sqrt{\alpha_i(1)}\bm{v}_i(1)$, where $\alpha_i(1)$ and $\alpha_i(1)$ are the exact Top-1 and -2 singular values and $\bm{v}_i(1)$ is the exact top singular vector. After $O(\frac{1}{\Delta}\log{\frac{n}{\epsilon}})$ steps, the local power method can achieve $\epsilon$-accuracy.
\end{theorem}
\begin{remark}
We defer the proof to the Appendix \ref{6}. According to Theorem 1, the normalized iterate $\frac{\mathbf{A}_{i}\mathbf{a}_{i}^{l}}{\|\mathbf{A}_{i}\mathbf{a}_{i}^{l}\|}$ is an $\epsilon$-accurate estimate of the top principal component. Hence, the total local power iterations complexity is $O(\frac{1}{\Delta}\log{\frac{n}{\epsilon}})$. \color{black} Our proof follows \cite{App}, which studies the centralized setting where all data are collected together and processed by a central machine. While in our case, there are two main differences: 1) our data is vertically distributed among clients and more importantly the raw data should not be exchanged in respect to the Federated Learning; 2) the global consensus of PCA/KPCA is reached by cooperated computations of all clients and the server (the server exists only for the master-slave network topology case). As a result, we have additional algorithmic challenges than [32] that mainly concerns computational efficiency only, such as the privacy challenge, communication efficiency challenge. \color{black}
\end{remark}

\section{Experimental Results}\label{4}
This section will empirically evaluate the proposed method. In the experiments, we utilized FIVE types of real-world datasets coming with distinct nature: 1) structured datasets from different domains \cite{ref23}; 2) medical image dataset \cite{ref25}; 3) face image dataset \cite{ref24}; 4) gait image dataset \cite{ref28}; 5) person re-identification image dataset \cite{ref30}. The feature and sample size information of all datasets are summarized in Table~\ref{t2}. In addition, illustrative images of the image dataset are shown in Figure~\ref{fig4}, correspondingly.

\begin{table}[t]
\begin{center}
\caption{Summary of Real-world \& Synthetic Datasets}
\label{t2}
\resizebox{0.50\textwidth}{!}{
    \begin{tabular}{l l c c}
        \toprule
        % \noalign{\smallskip}
        Datasets & \# Name & \# Features & \# Samples\\
        % \noalign{\smallskip}
        \midrule
        % \noalign{\smallskip}
        \multirow{8}{*}{Structured\footnote{http://archive.ics.uci.edu/ml/index.php}} & College & 15 & 777 \\
         & Vehicle & 18 & 846\\
         \cmidrule(lr){2-4}
         & PimaIndiansDiabetes & 33 & 351\\
         & GlaucomaM & 54 & 196\\
         & Sonar & 60 & 208\\
         & Musk & 166 & 476\\
         \cmidrule(lr){2-4}
         & Swarm & 2,400 & 2,000\\
         & TCGA & 20,500 & 4,000\\
        \midrule
        \midrule
        % \noalign{\smallskip}
        - & \# Distribution  & \# Features & \# Samples\\
        % \noalign{\smallskip}
        \midrule
        % \noalign{\smallskip}
        % \\ $X\sim \mathcal{N}(\mu, \sigma^{2})$
        \multirow{8}{*}{\tabincell{c}{Synthetic}} & \multirow{4}{*}{Single Gaussian} & 1,000 & 200 \\
         & & 5,000 & 1,000 \\
        & & 10,000 & 2,000\\
        & & 20,000 & 4,000\\
         \cmidrule(lr){2-4}
         & \multirow{4}{*}{\tabincell{c}{Mixture Gaussian}} & 1,000 & 200\\ %\\$X\sim \mathcal{N}(\mu, \sigma^{2})$
         & & 5,000 & 1,000 \\
         & & 10,000 & 2,000 \\
         & & 20,000 & 4,000 \\
        \midrule
        \midrule
         - & \# Image Size & \# Features & \# Samples\\
        \midrule
        % \midrule
        Face\footnote{http://vision.ucsd.edu/content/yale-face-database} & $100\times100$ & 10,000 & 225\\
        CUHK03\footnote{} & $128\times128$ & 16,384 & 14,097\\
        % \midrule
        CASIA\footnote{} & $256\times256$ & 65,536 & 240\\
        % \midrule
        DeepLesion\footnote{https://www.nih.gov/news-events/news-releases/nih-clinical-center-releases-dataset-32000-ct-images} & $512\times512$ & 262,144 & 32,000\\
        \bottomrule
    \end{tabular}}
\end{center}
\end{table}

\subsection{Experiment on Structured Dataset}\label{4.1}
\subsubsection{Semi-Synthetic Datasets}\label{4.1.1}
We use 8 structured datasets from different domains, which are the real data publicly available at the UCI Machine Learning Repository \cite{ref23}. we compare the communication cost and estimation error by the different settings of $\mathnormal{p}$ and $\mathnormal{l}$ based on the feature-wise setting. We change the number of clients $\mathnormal{p}$={2,3,5}, $\mathnormal{p}$={3,5,10}, and $\mathnormal{p}$={10,50,100}, and the local iterations $\mathnormal{l}$={3,5,10}, $\mathnormal{l}$={5,10,20}, $\mathnormal{l}$={30,50,100 and we set the number of communication period $\mathnormal{t}$=10. The number of each client's features are split and configured according to different settings of $\mathnormal{p}$.

\begin{figure}[t]
\centering
\includegraphics[width=0.85\columnwidth]{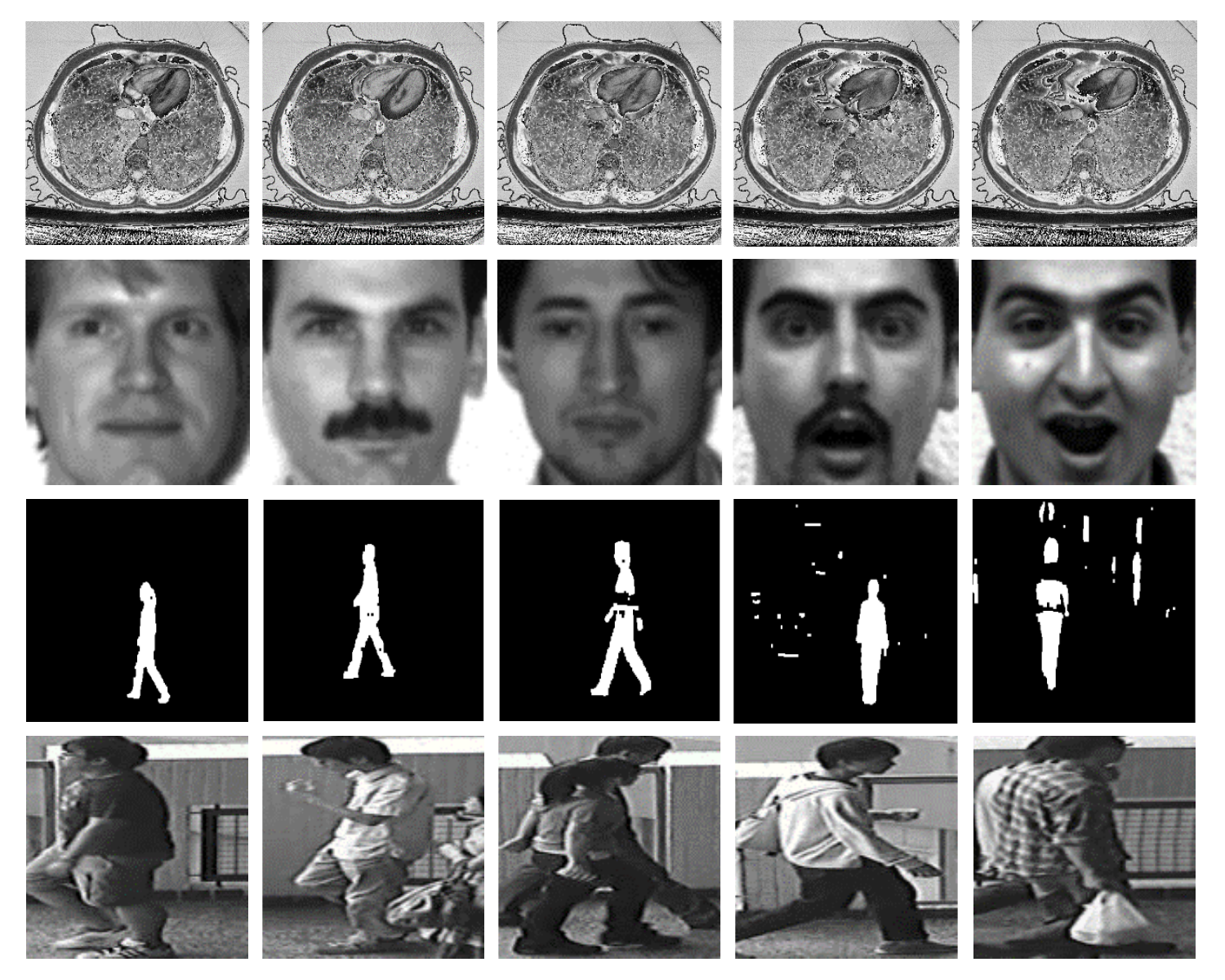}
\caption{The real-world datasets coming with distinct nature are from DeepLesion Dataset, YaleFace Dataset, CASIAGait Database, CUHK03 Database from top to bottom.}
\label{fig4}
\end{figure}

\subsubsection{Communication Cost and Estimation Error}\label{4.1.2}
In this part, we simulate the communication between the clients and the server on a single machine, where the CPU time are wall-clock time.
\begin{itemize}

\item\textbf{The effect of clients:}
In Figure \ref{example6}~(a) and \ref{example6}~(b), while we fix $\mathnormal{l}$=5, $\mathnormal{l}$=10 and $\mathnormal{l}$=100 and change the number of involved clients $\mathnormal{p}$={2,3,5}, $\mathnormal{p}$={3,5,10} and $\mathnormal{p}$={10,50,100}, respectively. Then, we plot the distance error between the global eigenvector and the federated eigenvector with the different clients $\mathnormal{p}$. In all experiments, compared with the un-communicated situation, the distance error after the communication has been significantly reduced, followed by a slight increase in some datasets, indicating that the multi-shot is not beneficial for all cases. It can be observed that the larger $\mathnormal{p}$ will lead to larger communication cost. % and the running time cost
 \item\textbf{The effect of local iterations:}
In Figure \ref{example6}~(c) and \ref{example6}~(d), while we fix $\mathnormal{p}$=3, $\mathnormal{p}$=5 and $\mathnormal{p}$=50 and change the number of local power iterations $\mathnormal{l}$={3,5,10}, $\mathnormal{l}$={5,10,20} and $\mathnormal{l}$={30,50,100}, respectively. \color{black} Most of the experimental results show that, after the communication, the distance error decreases and remains stable finally. However, some rise results also confirms that multi-shot communication do not benefit all datasets. \color{black}

 \item\textbf{The effect of warm-start power iterations:}
 In Figure \ref{example14}, we fix $\mathnormal{p}$={3,5,10} and local power iterations $\mathnormal{l}$=10. \color{black} After communications $\mathnormal{t}$=10, the experimental result shows that the larger p will get a significant result.\color{black}
 \item\textbf{The effect of $\eta$:}
In Figure \ref{example14}, we fix $\mathnormal{p}$={3,5,10} and local power iterations $\mathnormal{l}$=10. \color{black} After iterations $\mathnormal{t}$=10, the experimental result shows that the distance error further decreases and converges compared with not using the adjustment factor $\eta$.\color{black}
\color{black}
 \item\textbf{The result of vertical and horizontal FPCA:}
Figure \ref{example12_2} shows that the difference between vertical federated PCA and horizontal federated PCA.
\color{black}
\end{itemize}

\begin{figure*}[htbp]
\centering
\subfigure[]{\includegraphics[width=0.9\columnwidth]{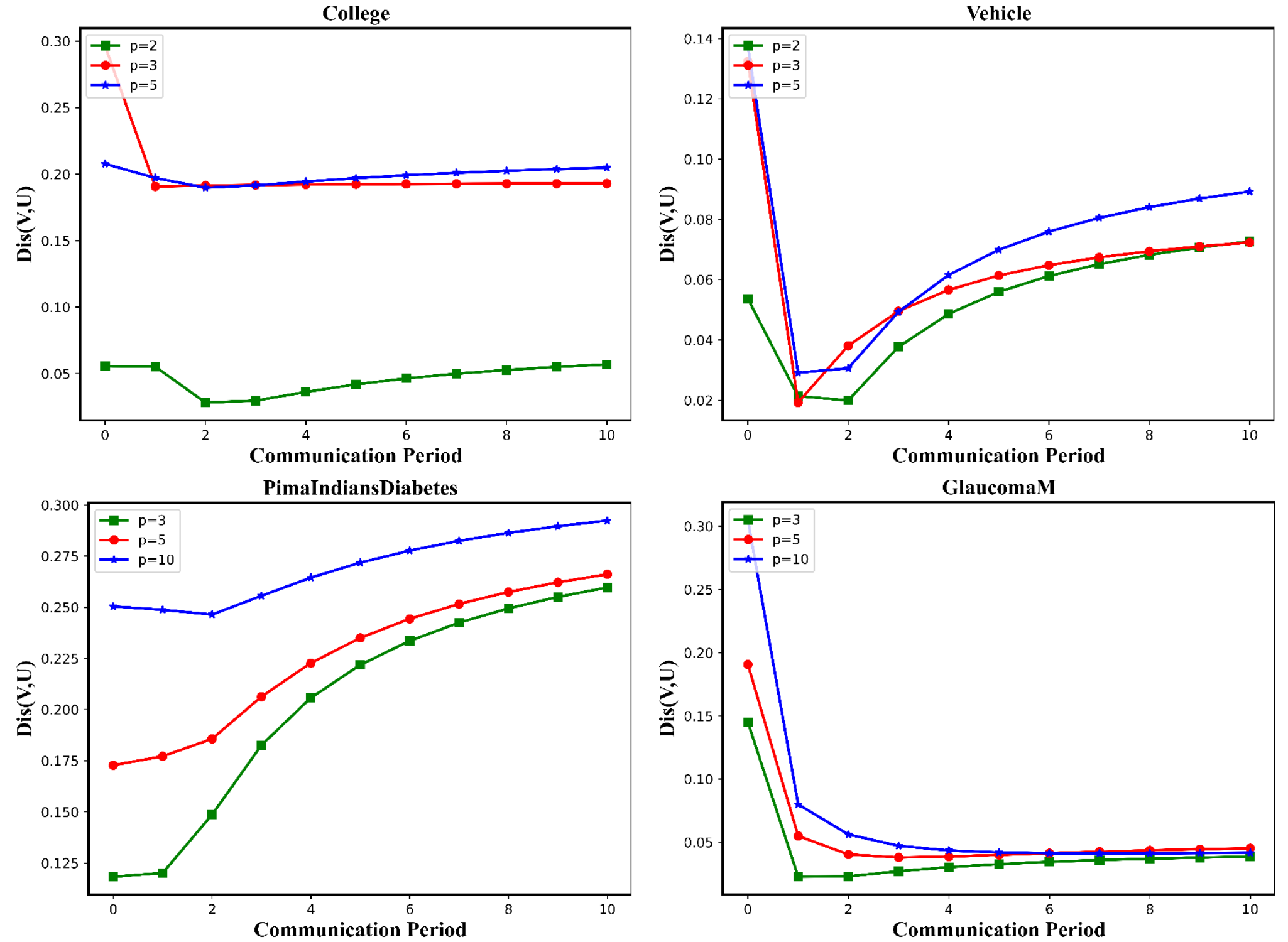}}\hspace{3mm}
\subfigure[]{\includegraphics[width=0.9\columnwidth]{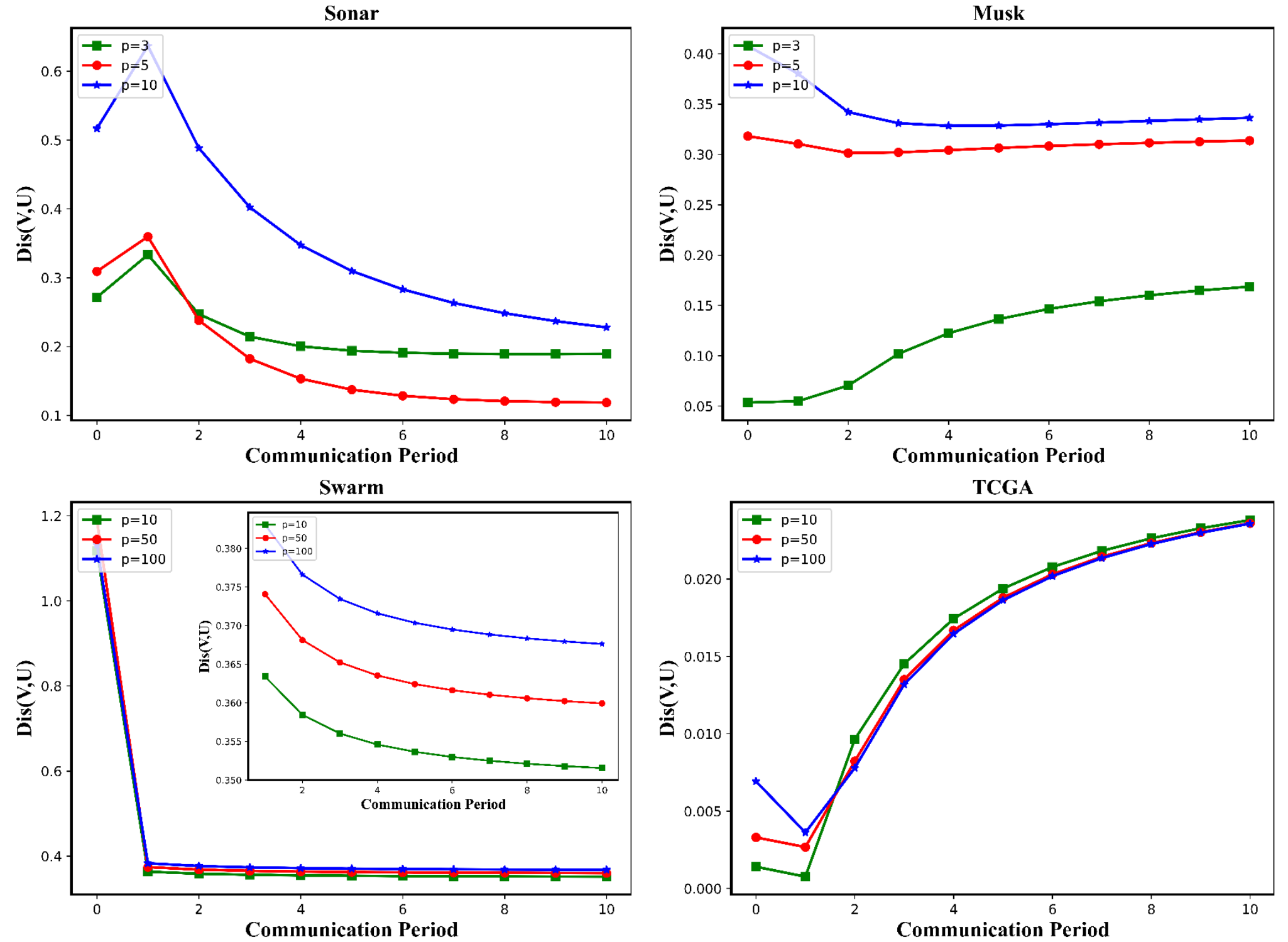}}\hspace{3mm}
\subfigure[]{\includegraphics[width=0.9\columnwidth]{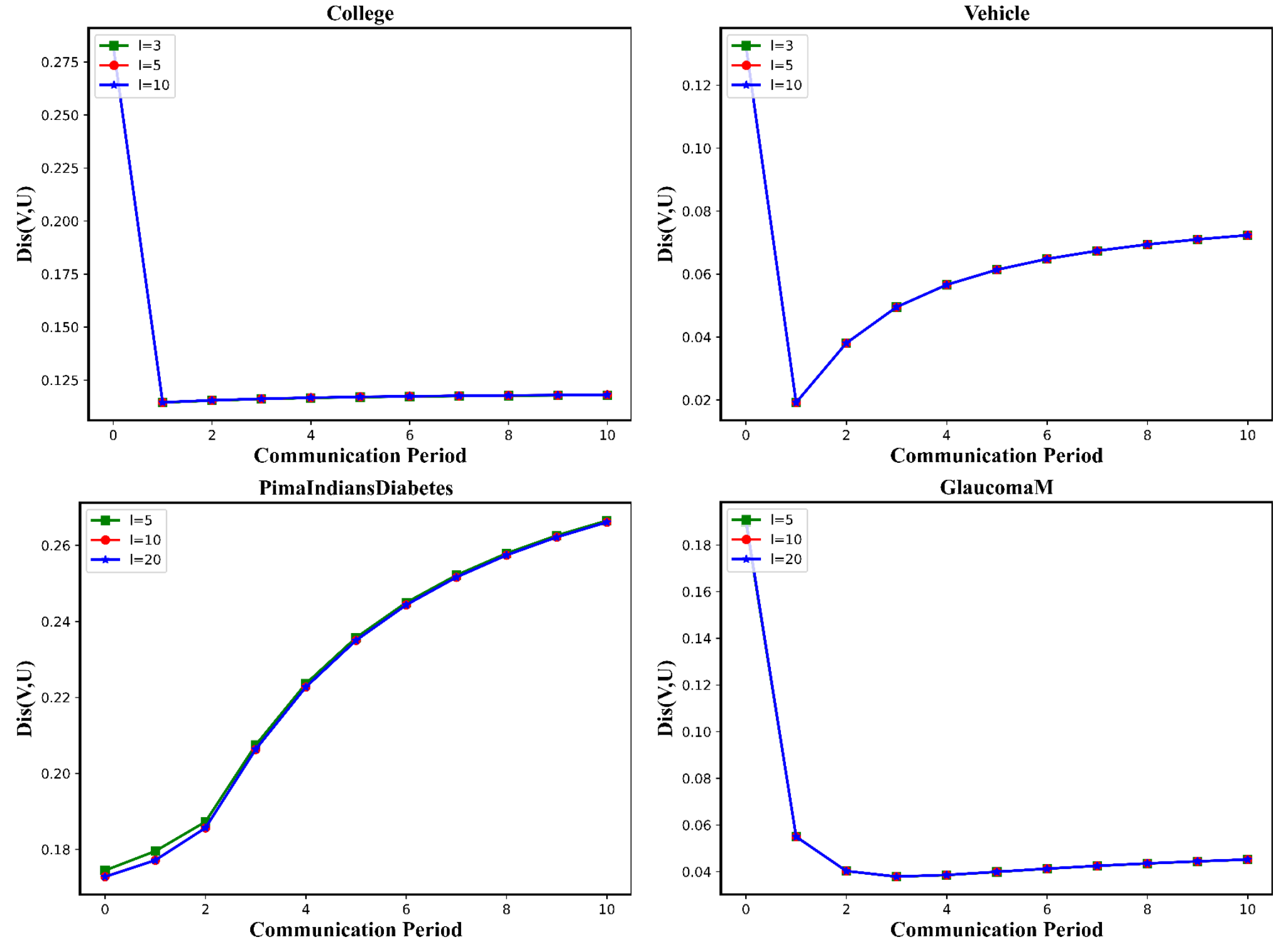}}\hspace{3mm}
\subfigure[]{\includegraphics[width=0.9\columnwidth]{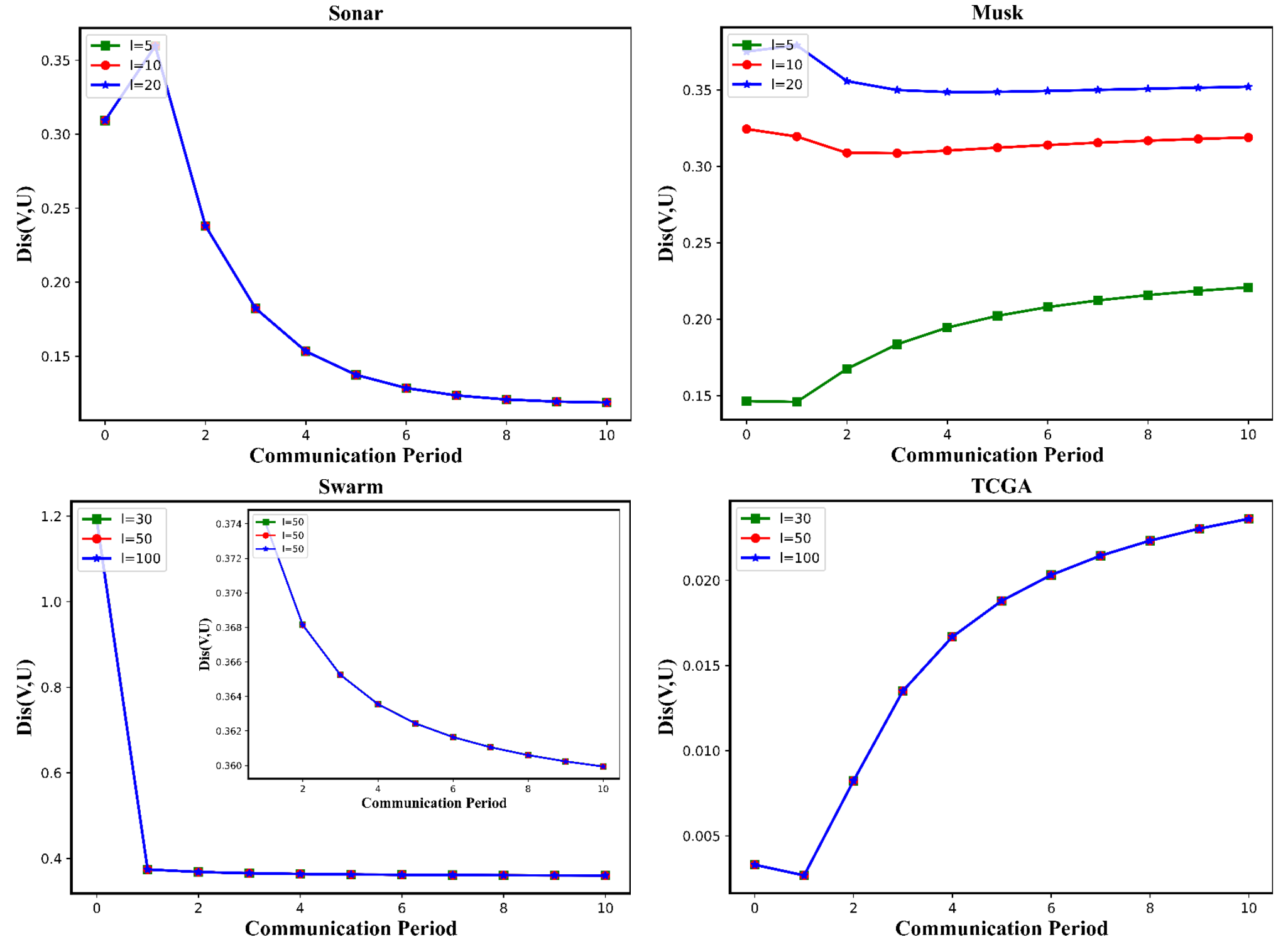}}
\caption{The results of VFedPCA on structured datasets with the effect of clients (a)(b) and local iterations (c)(d). (a) College, Vehicle, PimaIndiansDiabetes and GlaucomaM; For College and Vehicle, $\mathnormal{p}$=2,3,5, $\mathnormal{l}$=5; for PimaIndiansDiabetes and GlaucomaM, $\mathnormal{p}$=3,5,10, $\mathnormal{l}$=10; and (b) Sonar, Musk, and Swarm and TCGA; For Sonar and Musk, $\mathnormal{p}$=3,5,10, $\mathnormal{l}$=10; for Swarm and TCGA, $\mathnormal{p}$=10,50,100, $\mathnormal{l}$=100; and (c) College, Vehicle, PimaIndiansDiabetes and GlaucomaM; For College and Vehicle, $\mathnormal{p}$=3, $\mathnormal{l}$=3,5,10; for PimaIndiansDiabetes and GlaucomaM, $\mathnormal{p}$=5, $\mathnormal{l}$=5,10,20; and (d) Sonar, Musk, Swarm and TCGA. For Sonar, Musk, $\mathnormal{p}$=5, $\mathnormal{l}$=5,10,20; for Swarm and TCGA, $\mathnormal{p}$=50, $\mathnormal{l}$=30,50,100.}
\label{example6}
\end{figure*}

\begin{figure*}[htbp]
\centering
\subfigure[]{\includegraphics[width=0.9\columnwidth]{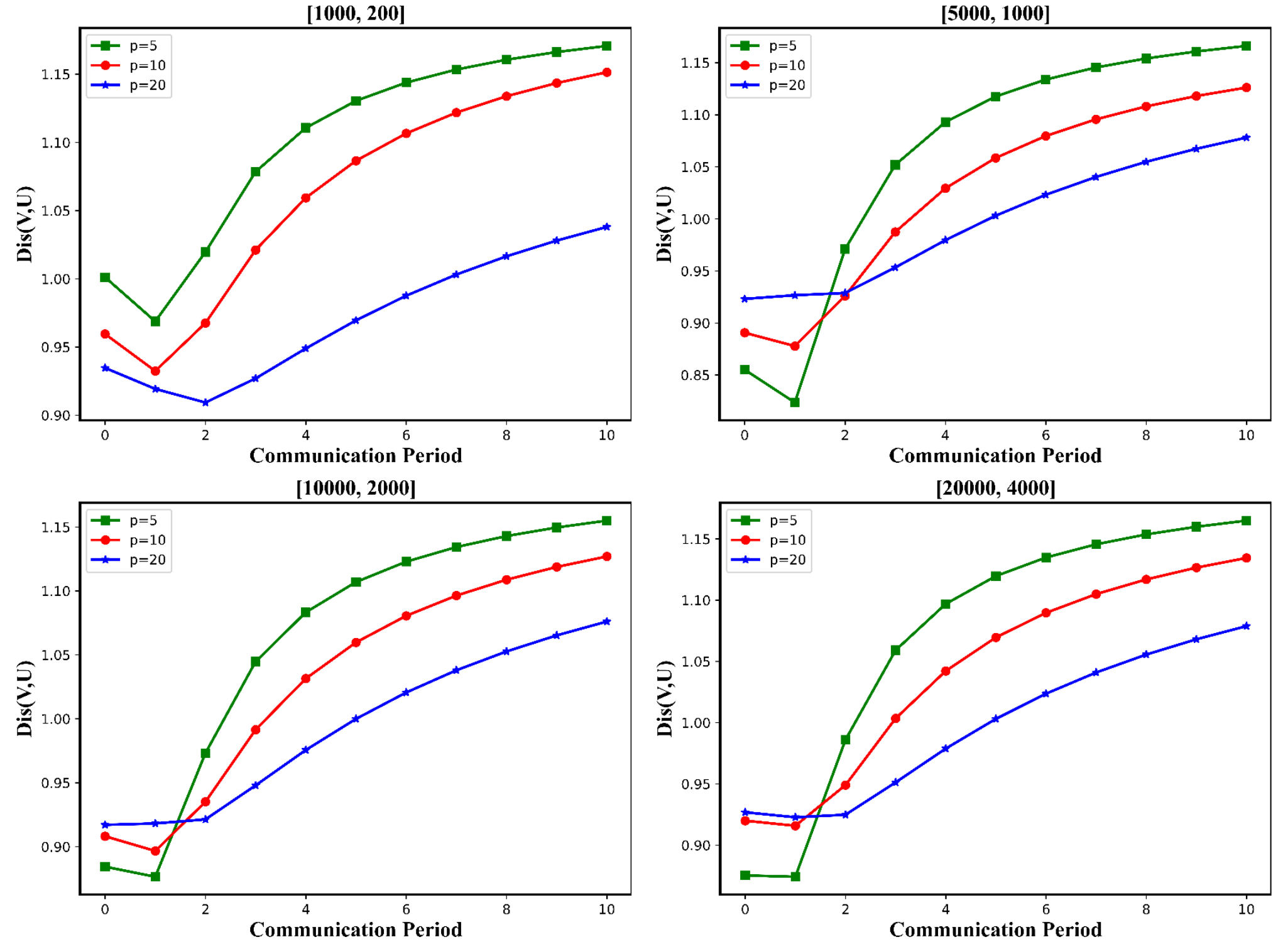}}\hspace{3mm}
\subfigure[]{\includegraphics[width=0.9\columnwidth]{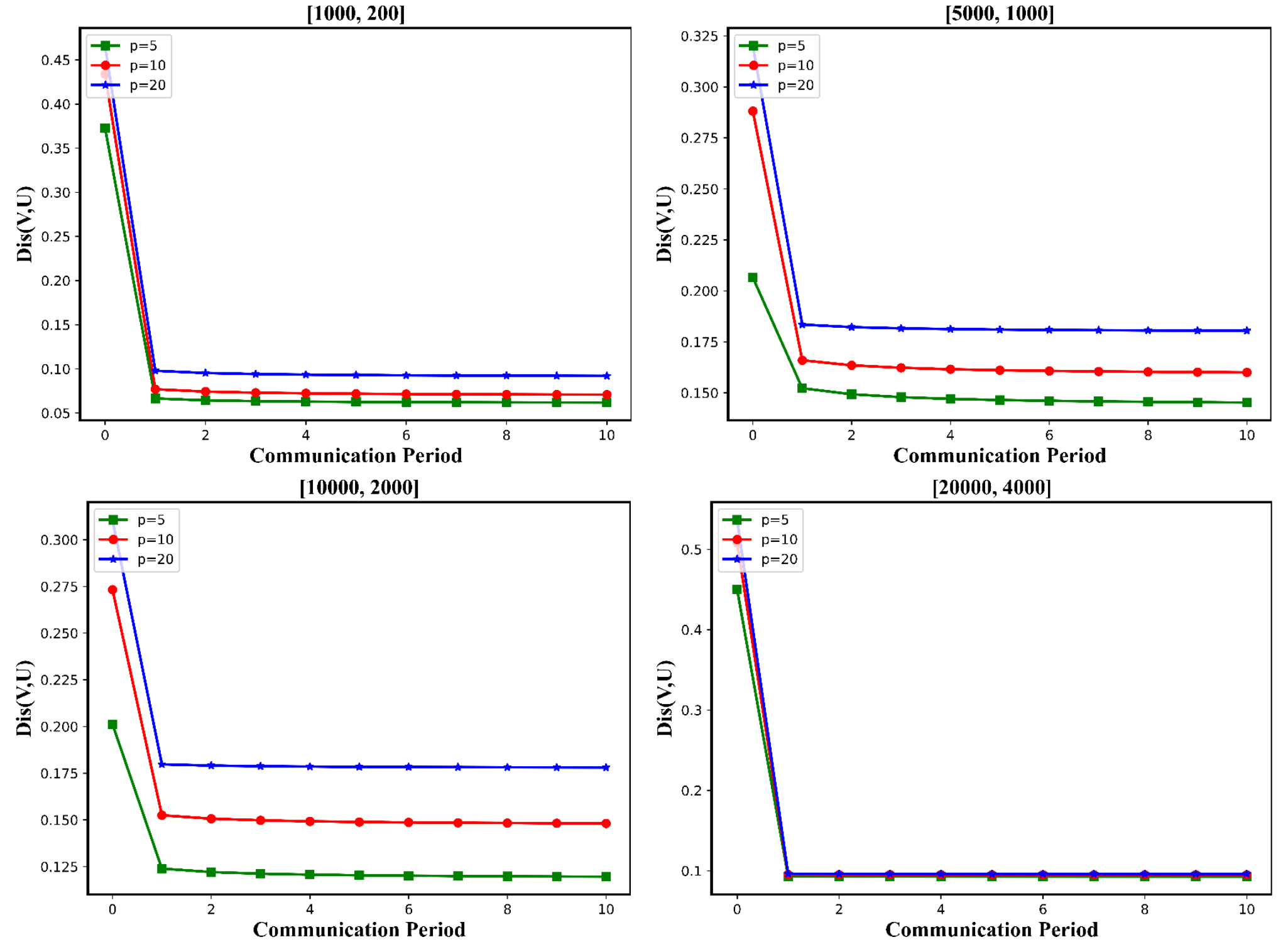}}\hspace{3mm}
\subfigure[]{\includegraphics[width=0.9\columnwidth]{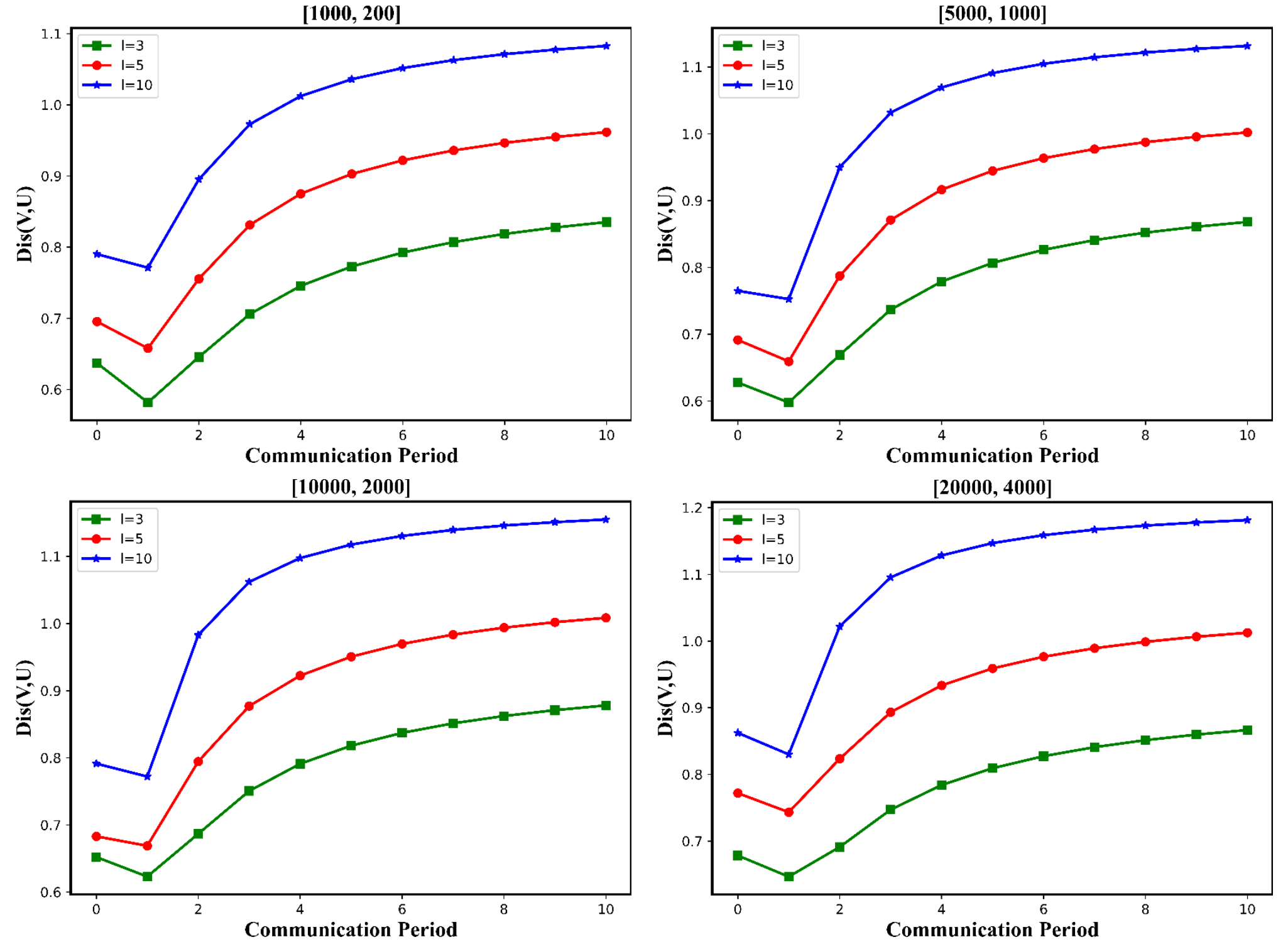}}\hspace{3mm}
\subfigure[]{\includegraphics[width=0.9\columnwidth]{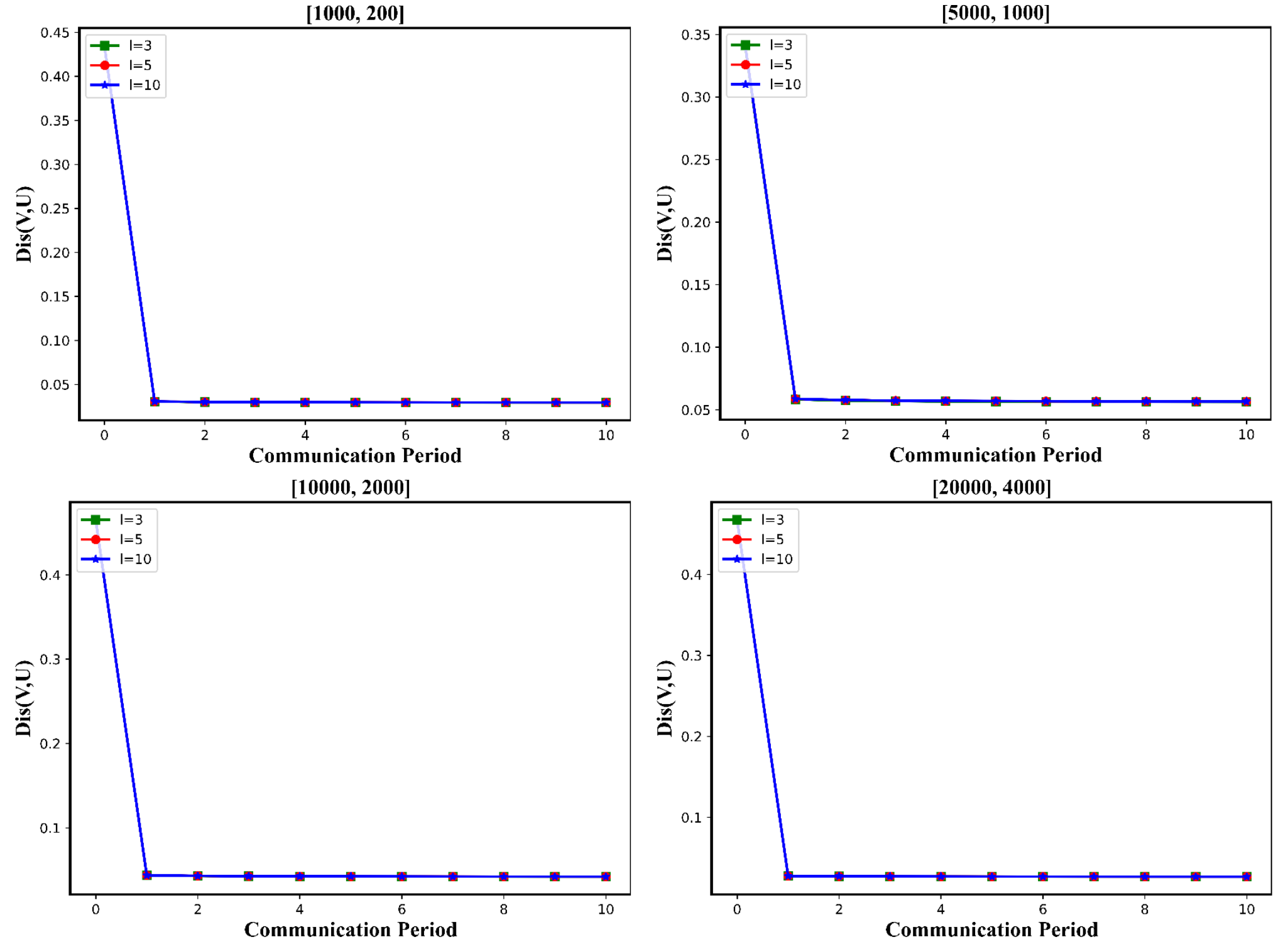}}
%\caption{The distance error(left) and the running time(right) of Algorithm 2 with respect to the number of parties when  \(\mathnormal{l}\)=10, \(\mathnormal{p}\)=3,5,10 on structured datasets: (a) College, (b) GlaucomaM, (c) PimaIndiansDiabetes, (d) Musk, (e) Vehicle, (f) Sonar; and \(\mathnormal{l}\)=100, \(\mathnormal{p}\)=10,50,100 on datasets (g) Swarm (h) TCGA.}
%\caption{The distance error (left side) and the running time (right side) of Algorithm \ref{a1} with respect to the number of clients on synthetic datasets: (a) [1000,200] and [5000,1000]; and (b) [10000,2000], [20000,4000]. For (a) and (b): $\mathnormal{p}$=5,10,20, $\mathnormal{l}$=10.}
\caption{The results of VFedPCA on synthetic datasets from two distribution (Single and Mixture Gaussian) with the effect of clients (a)(b) and local iterations (c)(d), by the different settings of [features, samples] = [1000,200], [5000,1000], [10000,2000] and [20000,4000]. For (a) $\mathnormal{p}$=5,10,20, $\mathnormal{l}$=10; for (b) $\mathnormal{p}$=5,10,20, $\mathnormal{l}$=10; for (c) $\mathnormal{p}$=3, $\mathnormal{l}$=3,5,10; for (d) $\mathnormal{p}$=3, $\mathnormal{l}$=3,5,10.}
\label{example10}
\end{figure*}

\begin{figure}[t]
\centering
\includegraphics[width=0.9\columnwidth]{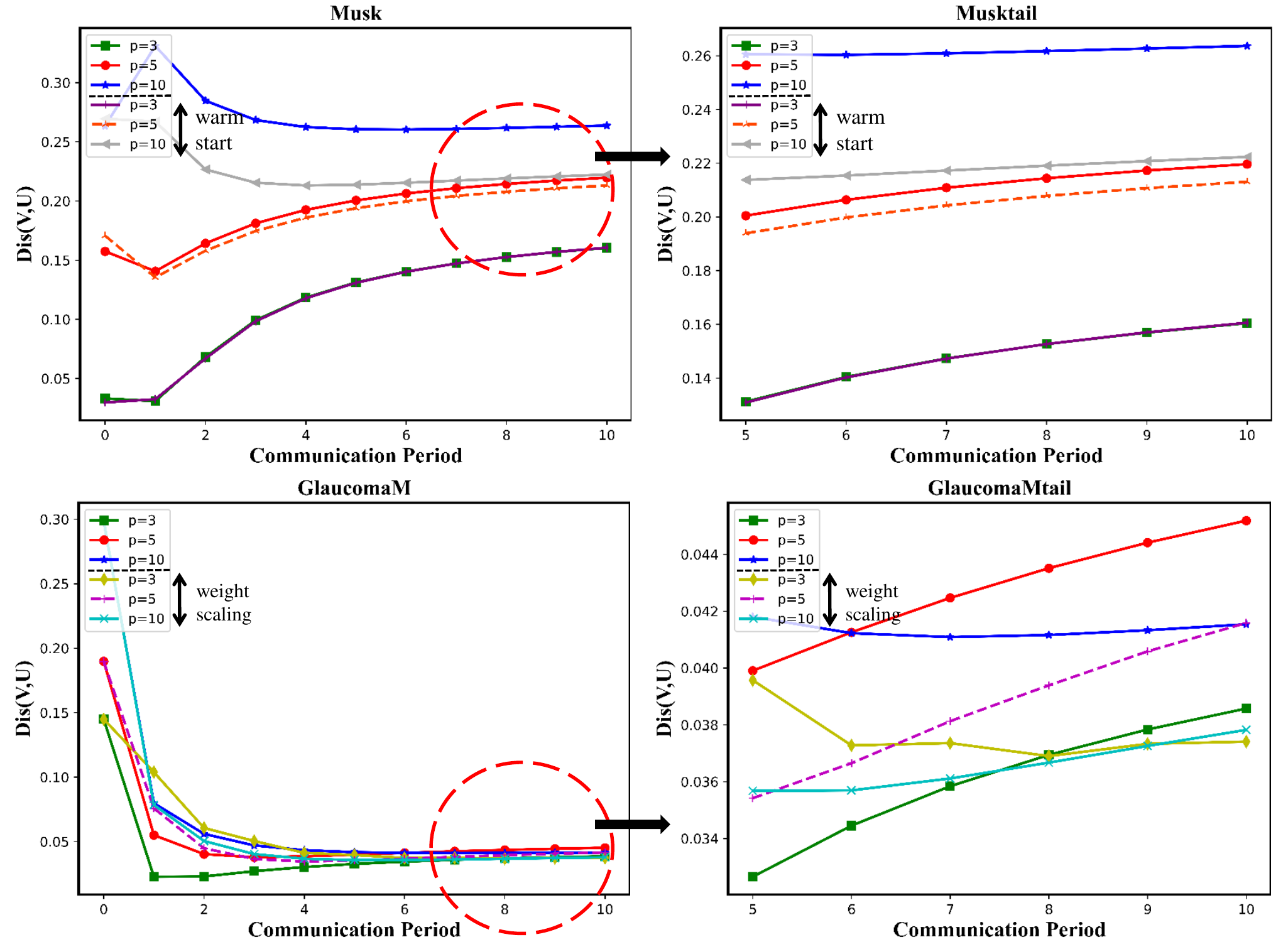}
%\caption{The distance error of Algorithm \ref{a3} (upper) with respect to the number of clients when $\mathnormal{p}$=3,5,10, $\mathnormal{l}$=10 on structured datasets: Musk (upper) and Algorithm \ref{a4} with respect to the number of clients when $\mathnormal{p}$=3,5,10, $\mathnormal{l}$=10 on structured datasets: GlaucomaM (lower).}
\caption{The results of Algorithm \ref{a3} (upper) and Algorithm \ref{a4} (lower) on structured datasets: Musk (upper) and GlaucomaM (lower).}
\label{example14}
\end{figure}

\begin{figure}[t]
\centering
\includegraphics[width=0.9\columnwidth]{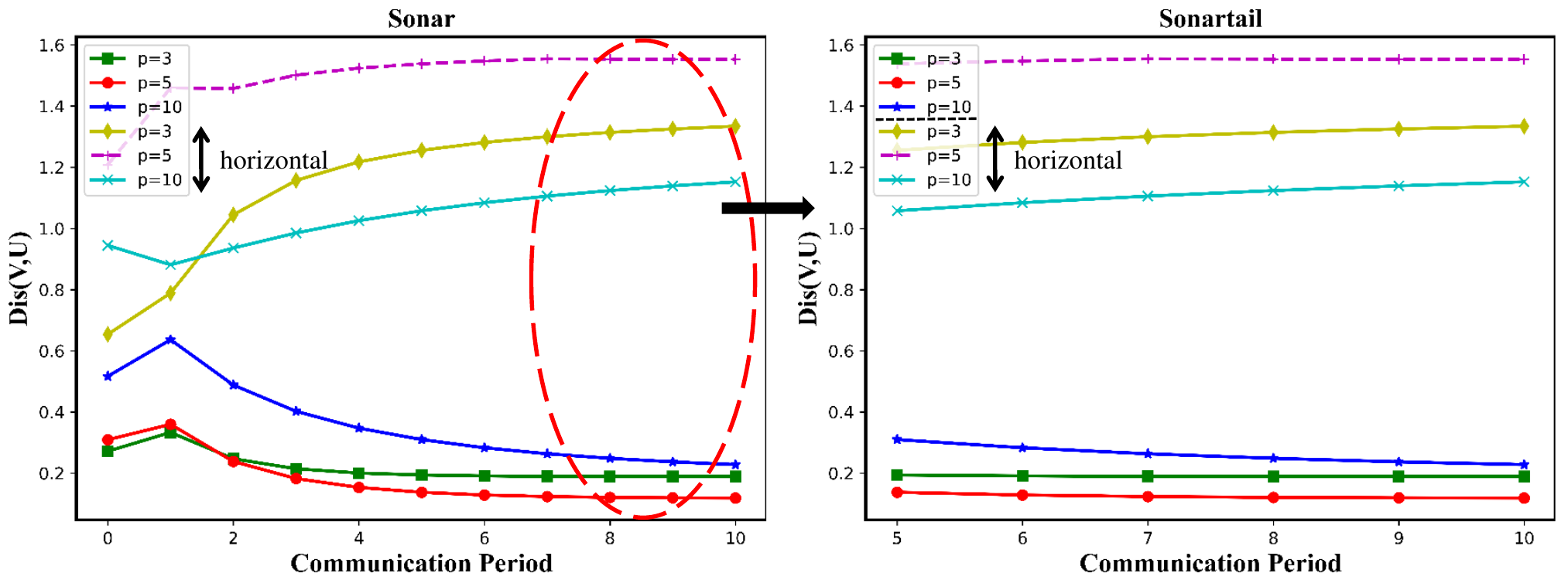}
%\caption{The distance error of vertical and horizontal Federated PCA with respect to the number of clients when $\mathnormal{p}$=3,5,10, $\mathnormal{l}$=10 on structured datasets: Sonar.}
\caption{The results of vertical and horizontal Federated PCA on structured datasets: Sonar.}
\label{example12_2}
\end{figure}

\subsection{Experiment on Synthetic Dataset}

\subsubsection{Synthetic Datasets}\label{4.1.1}
We use synthetic datasets from two distribution: single Gaussian and mixture Gaussian by the different settings of [features, samples] = [1000,200], [5000,1000], [10000,2000] and [20000,4000]; with the number of clients $\mathnormal{p}$={5,10,20}, and the local iterations $\mathnormal{l}$={5,10,20}. \color{black} In practical scenarios, each dimension of features has its own numerical distribution and attributes. Following the practical pf synthetic data generation in FL literature like \cite{guo2021privacy}, we employ Gaussian distribution $X\sim \mathcal{N}(\mu, \sigma^{2})$ with different $\mu$ and $\sigma$ in each dimension of features to generate synthetic datasets, termed as mixture of Gaussian distribution datasets. In particularly, we also consider all features obey the same Gaussian distribution, termed as  single Gaussian distribution datasets. In our experiments, the $\mu$ and $\sigma$ are randomly sampled from $[0, 1, 2, ..., m]$, and each pair of $(\mu_i, \sigma_i)$ represents the numerical distribution of feature $i$-th of $m$. Then, we randomly sample $n$ points from $x \in (\mu_i-3\sigma_i, \mu_i+3\sigma_i)$ interval, and calculate the corresponding $y=\frac{1}{\sqrt{2\pi}\sigma_i}\exp{\left( -\frac{(x-\mu_i)^2}{2\sigma_i^2}\right)}\in\mathbb{R}^{n\times1}, i\in m$ as the synthetic value. For single Gaussian dataset, the same synthetic strategy is applied, while using fixed $\mu=0$ and $\sigma=1$. In this way, we can generate the synthetic dataset $X\in\mathbb{R}^{n \times m}$\color{black}.
\color{black}
\subsubsection{Communication Cost and Estimation Error}\label{4.1.2}
In this part, we simulate the communication between the clients and the server on a single machine, where the CPU time are wall-clock time.
\begin{itemize}
\item\textbf{The effect of clients:}
In Figure \ref{example10}~(a) and \ref{example10}~(b), we fix $\mathnormal{l}$=10, while change the number of involved clients $\mathnormal{p}$={5,10,20}. Figure \ref{example10}~(a) shows that one-shot is more effective. Figure \ref{example10}~(b) shows that the distance error eventually decreases and converges after the communication.
 \item\textbf{The effect of local iterations:}
In Figure \ref{example10}~(c) and \ref{example10}~(d), we fix $\mathnormal{p}$=3, while change the number of local power iterations $\mathnormal{l}$={3,5,10}. Figure \ref{example10}~(c) shows that one-shot is more effective. Figure \ref{example10}~(d) shows that the distance error eventually decreases and converges after the communication.
\end{itemize}
\color{black}
% \begin{table}[t]
% \begin{center}
% \caption{Summary of Synthetic Datasets}
% \label{t3}
% \begin{tabular}{l l c c}
%     \toprule
%     % \noalign{\smallskip}
%     Distribution & \# Num & \# Features & \# Samples\\
%     % \noalign{\smallskip}
%     \midrule
%     % \noalign{\smallskip}
%     \multirow{8}{*}{\tabincell{c}{$X\sim \mathcal{N}(\mu, \sigma^{2})$}} & \multirow{4}{*}{Single} & 1,000 & 200 \\
%      & & 5,000 & 1,000 \\
%      & & 10,000 & 2,000\\
%      & & 20,000 & 4,000\\
%      \cmidrule(lr){2-4}
%      & \multirow{4}{*}{Mixture} & 1,000 & 200\\
%      & & 5,000 & 1,000 \\
%      & & 10,000 & 2,000 \\
%      & & 20,000 & 4,000 \\
%     \bottomrule
% \end{tabular}
% \end{center}
% \end{table}

\subsection{Case Studies}\label{4.2}
\subsubsection{Medical Image Dataset}\label{4.2.1}
The DeepLesion dataset \cite{ref25} is a CT slices collection from 4427 unique patients, which contains a variety of lesions (e.g., lung nodules, liver lesions). We select 100 samples from the dataset and each image is normalized to an 512\(\times\)512 gray image. We set $\mathnormal{p}$=10, 16, 8 clients, then re-splitted the features ($\mathnormal{d}$=10000, 65536, 262144), respectively, and assign them equally to each client. We use the common clustering method of k-means.

\subsubsection{Yale Face Dataset}\label{4.2.2}
We use the Yale Face Dataset \cite{ref24}, which contains 165 grayscale images of 15 subjects. Each subject configures 11 different facial expressions and $\mathnormal{n}$=15 samples for each facial expression, where each face image is normalized to an 100\(\times\)100 gray image. We set $\mathnormal{k}$=15 in the k-means clustering.

\subsubsection{Gait Estimation}\label{4.2.3}
The CASIA is a gait database \cite{ref28} for gait recognition, including 20 persons. It comes with 4 sequences for each of the three directions and $\mathnormal{n}$=20 samples for each direction, where each image is resized to the 256\(\times\)256 scale. We set $\mathnormal{k}$=10 in the k-means clustering.

\subsubsection{Person Re-Identification}\label{4.2.4}
The CUHK03 \cite{ref30} consists of 14,097 images of 1,467 different identities, where 6 campus cameras are deployed for image collection and each identity is captured by 2 campus cameras, each image is resized to the 128\(\times\)128 scale.

\begin{figure}[t]
\centering
\subfigure[]{
\includegraphics[width=0.85\columnwidth]{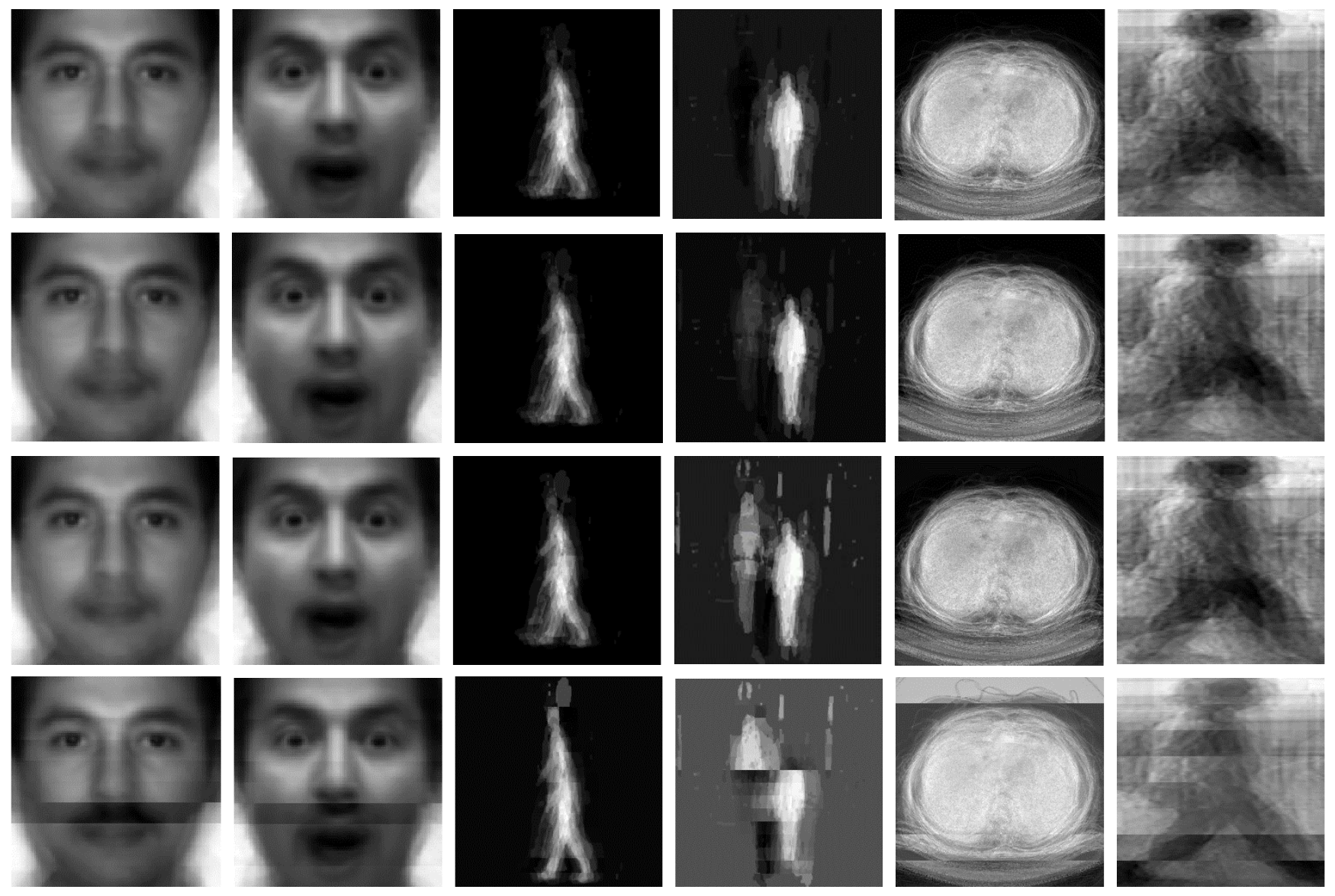}
}
\subfigure[]{
\includegraphics[width=0.85\columnwidth]{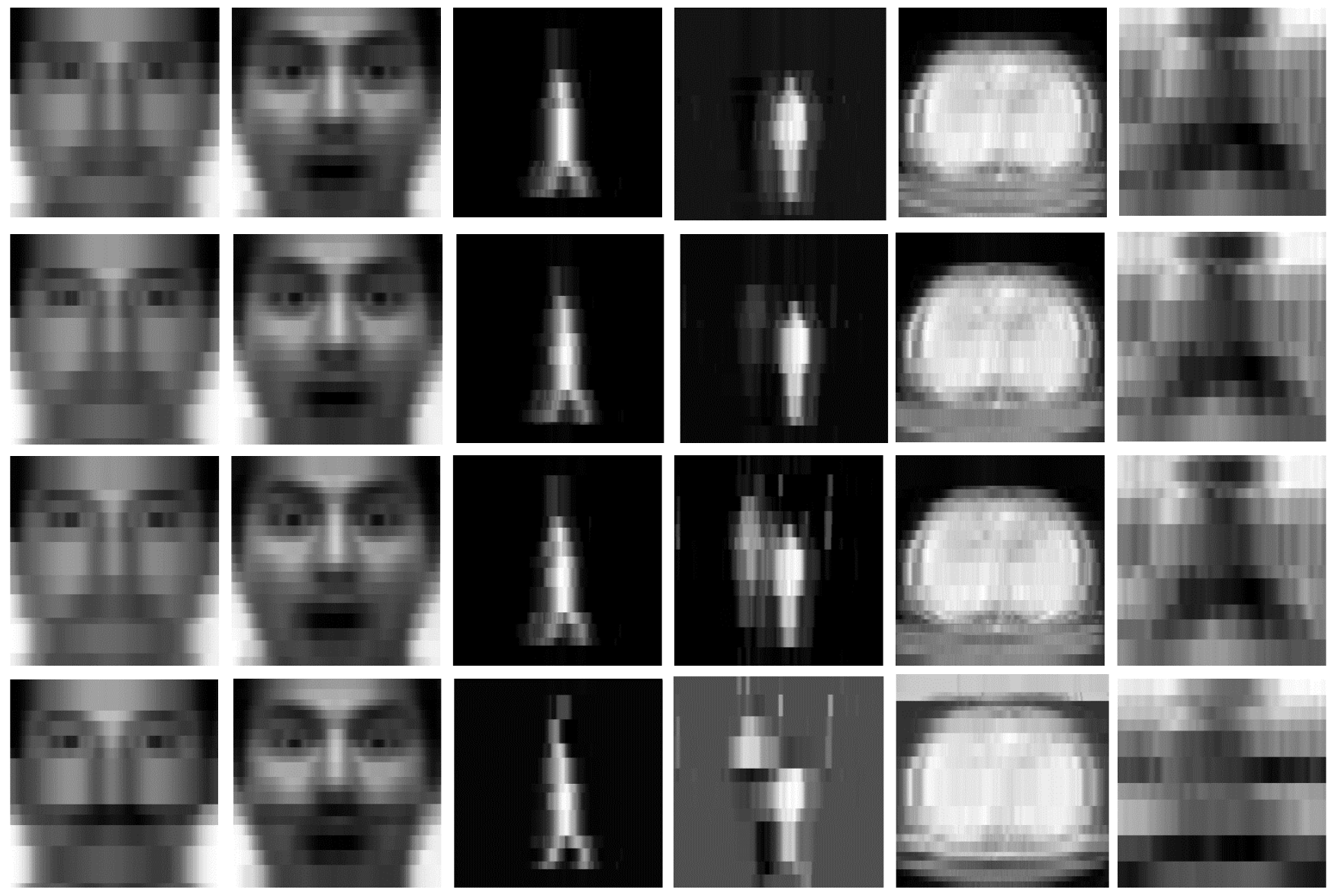}
}
%\caption{The final results of comparative experiment on image datasets: YaleFace (center-light and surprised), CasiaGait (sequence 1 and 10) and DeepLesion with algorithms: (a) PCA on the un-split data, VFedPCA on the split data, VFedAvgPCA on the split data, PCA on the isolated data, (d) the un-split data, (e) the federated data, (f) the isolated data.}
%\caption{The results of comparative experiment on image datasets: YaleFace (center-light and surprised), CasiaGait (sequence 1 and 10), DeepLesion and CUHK03 (from left to right) with algorithms: (a) PCA on the overall data, VFedPCA on the isolated data, VFedAvgPCA on the isolated data, PCA on the isolated data (from top to bottom). Image segmentation (k=10) results: (b) PCA on the overall data, VFedPCA on the isolated data, VFedAvgPCA on the isolated data, PCA on the isolated data (from top to bottom).}
\caption{The results of PCA on the un-splitted data, VFedPCA on the isolated data, VFedAvgPCA (Without Weight Scaling Method) on the isolated data, PCA on the isolated data (from top to bottom). For (b), after image segmentation respectively}
\label{fig:example4}
\end{figure}
\begin{figure}[t]
\centering
\includegraphics[width=0.85\columnwidth]{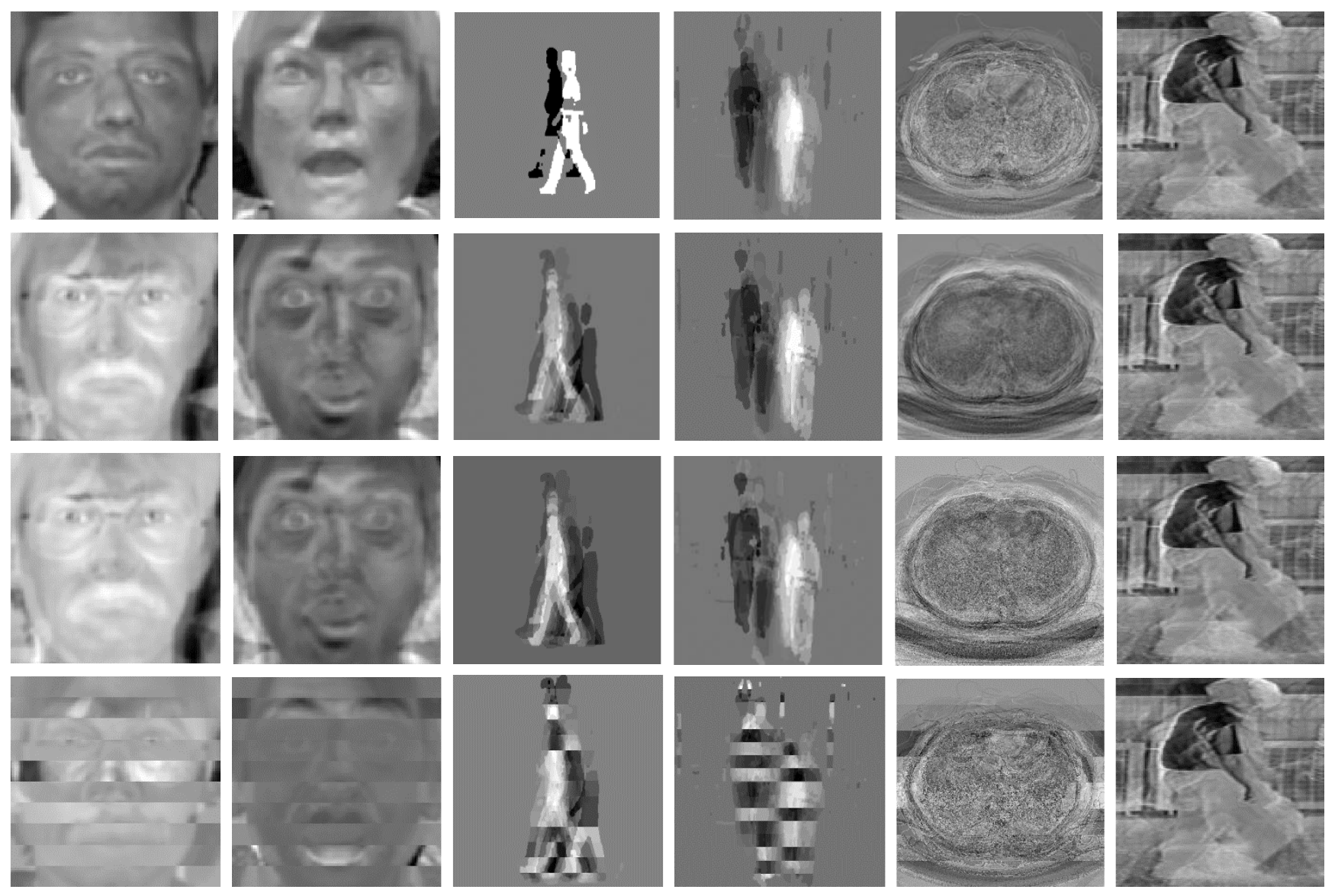}
%\caption{The results of comparative experiment on image datasets: YaleFace (center-light and surprised), CasiaGait (sequence 1 and sequence 10), DeepLesion and CUHK03 (from left to right) : AKPCA on the un-splitted data, VFedAKPCA on the isolated data, VFedAvgAKPCA on the isolated data, AKPCA on the isolated data (from top to bottom).}
%\caption{AKPCA vs. VFedAKPCA vs VFedAvgAKPCA on image datasets: YaleFace (center-light and surprised), CasiaGait (sequence 1 and 10), DeepLesion and CUHK03: AKPCA on the un-split data, VFedAKPCA on the split data, VFedAvgAKPCA on the split data, AKPCA on the isolated data.}
\caption{The results of AKPCA on the un-splitted data, VFedAKPCA on the isolated data, VFedAvgAKPCA (Without Weight Scaling Method) on the isolated data, AKPCA on the isolated data (from top to bottom).}
\label{fig:example6}
\end{figure}
\begin{figure}[t]
\centering
\subfigure[]{
\includegraphics[width=0.85\columnwidth]{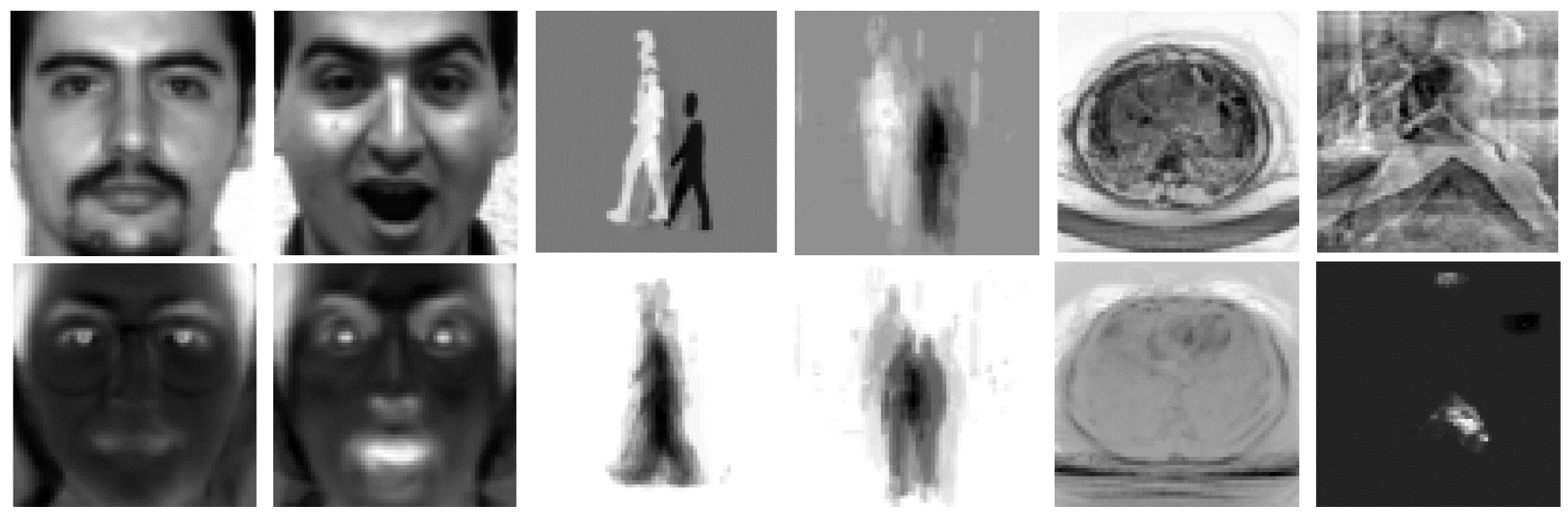}
}
\subfigure[]{
\includegraphics[width=0.85\columnwidth]{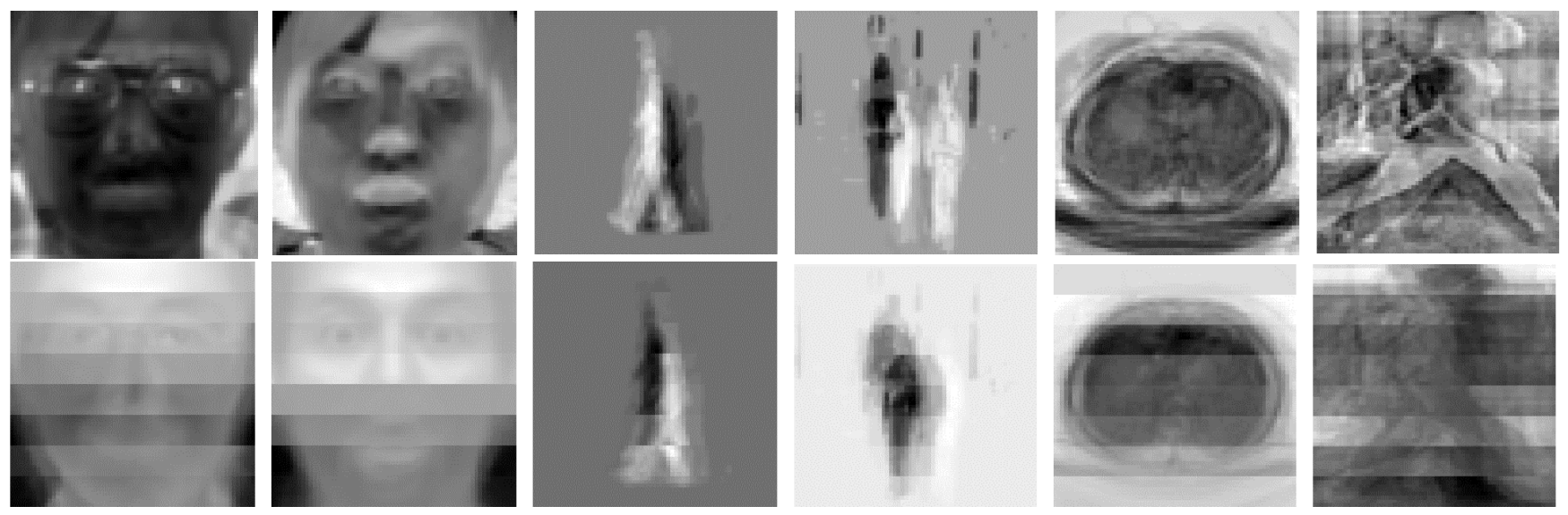}
}
\subfigure[]{
\includegraphics[width=0.85\columnwidth]{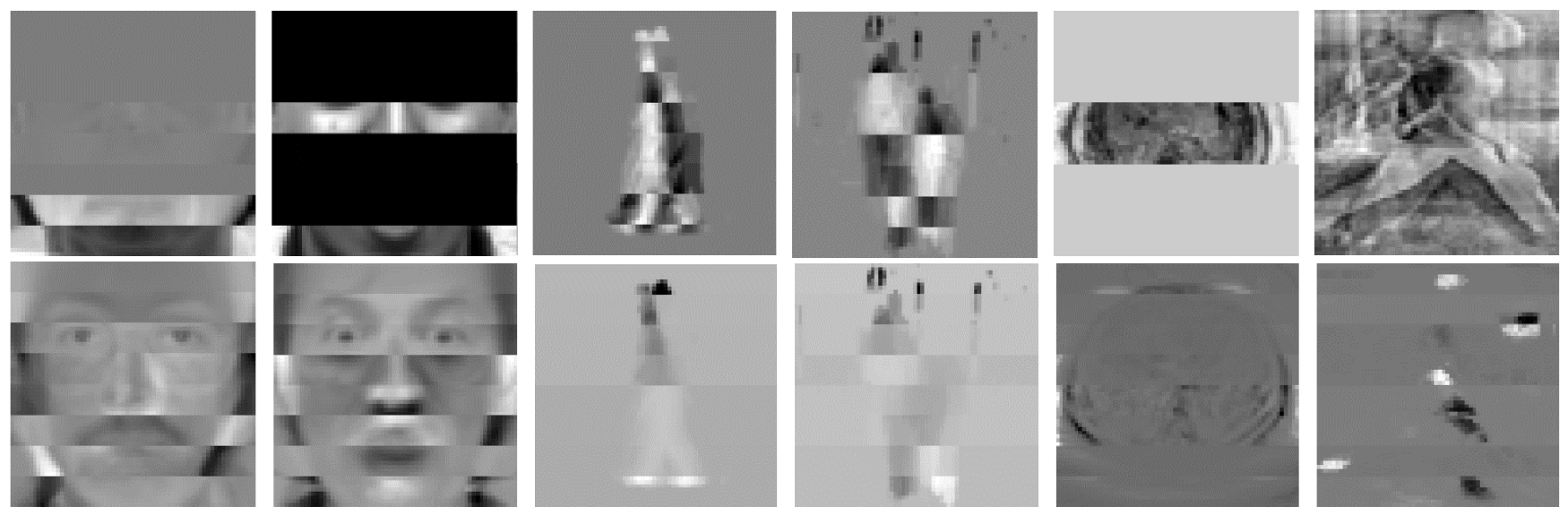}
}
%\caption{The results of comparative experiment on image datasets: YaleFace (center-light and surprised), CasiaGait (sequence 1 and 10), DeepLesion and CUHK03 (from left to right) with algorithms: (a) AKPCA (upper) and KPCA (lower) on the overall data, (b) VFedAKPCA (upper) and VFedKPCA (lower) on the isolated data, (c) AKPCA (upper) and KPCA (lower) on the isolated data.}
\caption{The results of (a) AKPCA/KPCA on the un-splitted data, (b) VFedAKPCA/VFedKPCA on the isolated data, (c) AKPCA/KPCA on the isolated data.}
\label{fig:example9}
\end{figure}

\begin{figure}[t]
\centering
\includegraphics[width=0.85\columnwidth]{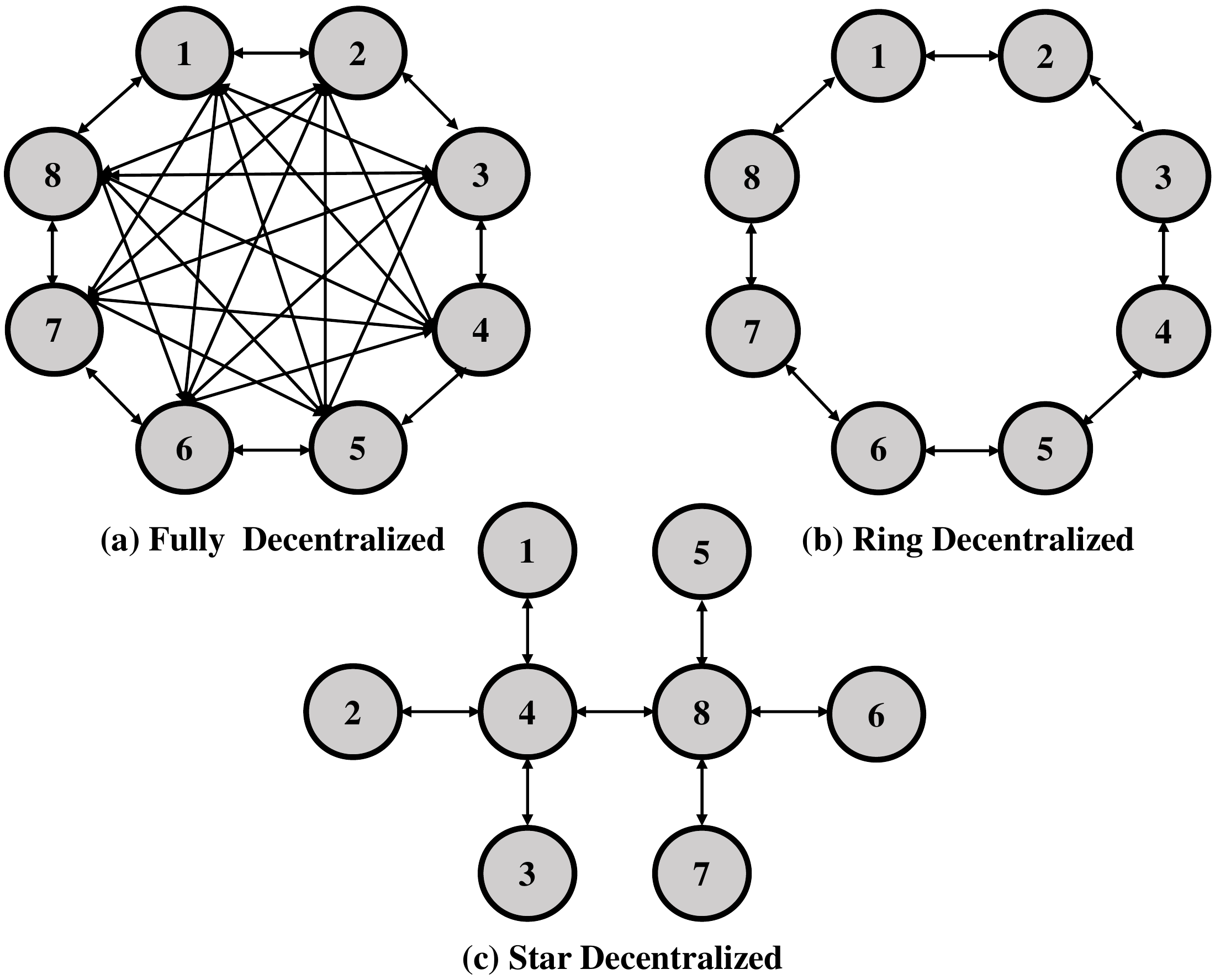}
\caption{The three different types of decentralized topology.}
\label{fig:example8-1}
\end{figure}

\color{black}
\subsubsection{Comparative Results}\label{4.2.5}
%We first perform PCA on image dataset. Figure~\ref{fig:example4}~(a)(b)(c) show that the final image after the proposed vertical federated PCA is almost the same as the final image given by centralized approach. Then, to further verify that the final image result extracted more features, we also performed the clustering experiment on the image dataset. In contrast to the settings where each client uses the standard PCA method independently, Figure~\ref{fig:example6}~(d)(e)(f) show the clustering result after using the federated PCA method, which also shows a different effect from the isolated PCA. Our federated counterpart will help each client to further outperform local training tasks. In Figure~\ref{fig:example6}~(a)(b)(c), compared with the centralized AKPCA,  our VFedAKPCA can extract more details of the image. PCA mainly focuses on extracting the overall structure information of the image.
In this part, to validate our proposed methods, we conduct comparative experiments on the above case studies.
\begin{itemize}[leftmargin=*]
\item\textbf{The evaluation of federated learning:}
First, we evaluate the ability of the VFedPCA and vertically federated average PCA (VFedAvgPCA) on different case studies compared with PCA on the overall dataset and PCA on the isolated dataset. Figure \ref{fig:example4}~(a) shows that the effect after using VFedPCA is significantly better than the case where using VFedAvgPCA, and using PCA on the isolated dataset, and the final image obtained is almost the same as the final image given by PCA on the overall dataset. Then, to further verify that the final image result extracted more features, we also performed the clustering experiment. In contrast to the settings where each client using the local power iteration (PCA) method independently, Figure~\ref{fig:example4}~(b) shows the clustering result after using the federated PCA method, which also shows a different effect from the isolated PCA. Our federated counterpart will help each client to further outperform local training tasks. In Figure \ref{fig:example6}, we evaluate the ability of the VFedAKPCA and VFedAvgAKPCA compared with AKPCA on the overall dataset and isolated dataset. It is obvious that the AKPCA focuses on extracting interesting nonlinear structures in images, it performs well in capturing details of
image details, although this may also ignore the overall structure. Figure \ref{fig:example9} shows that the difference between AKPCA and KPCA, we find that the AKPCA has shown good performance in terms of final effect and time complexity under the vertical data partition.

\begin{figure}[htbp]
\centering
\subfigure[]{
\includegraphics[width=0.85\columnwidth]{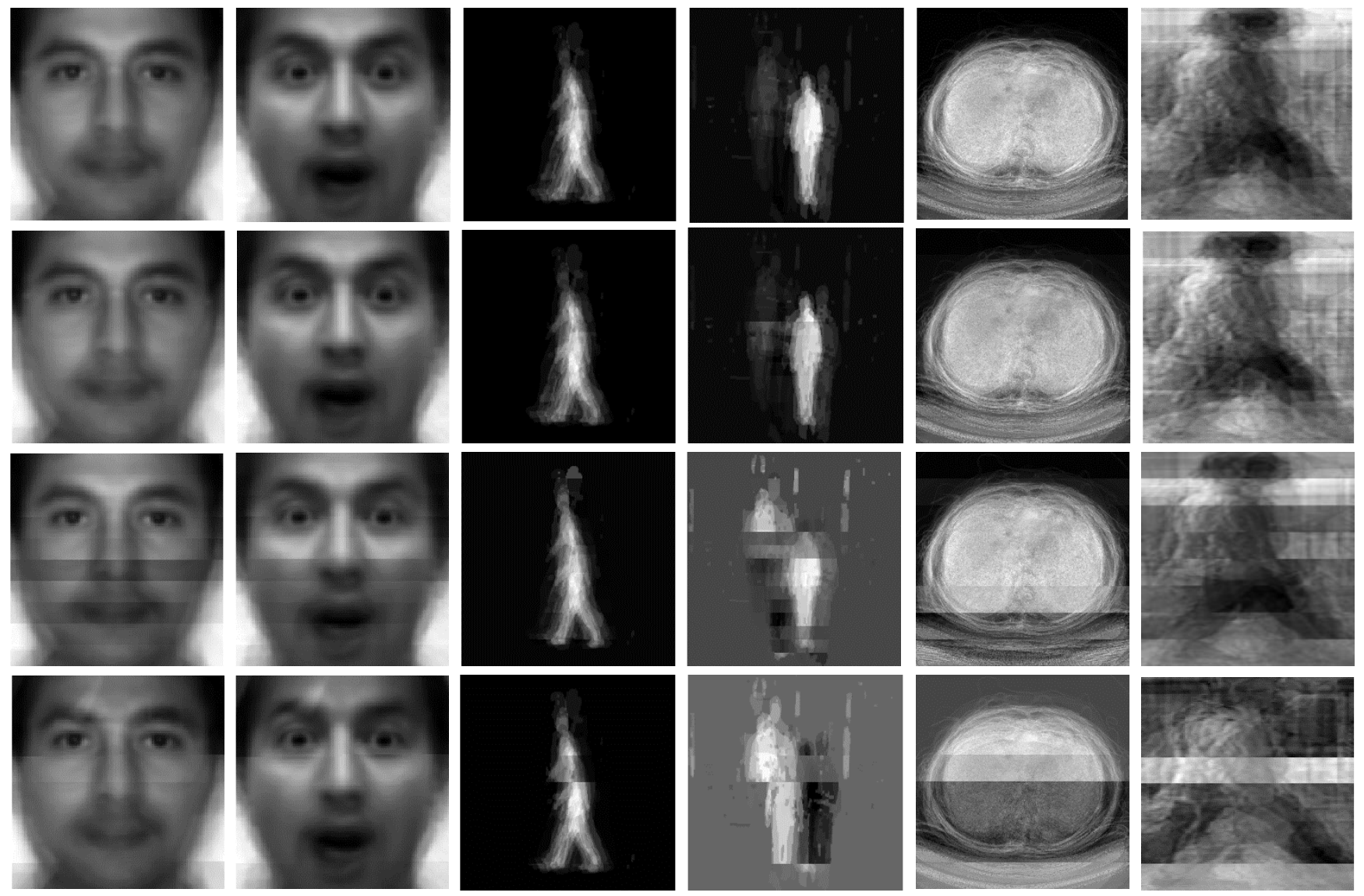}
}
\subfigure[]{
\includegraphics[width=0.85\columnwidth]{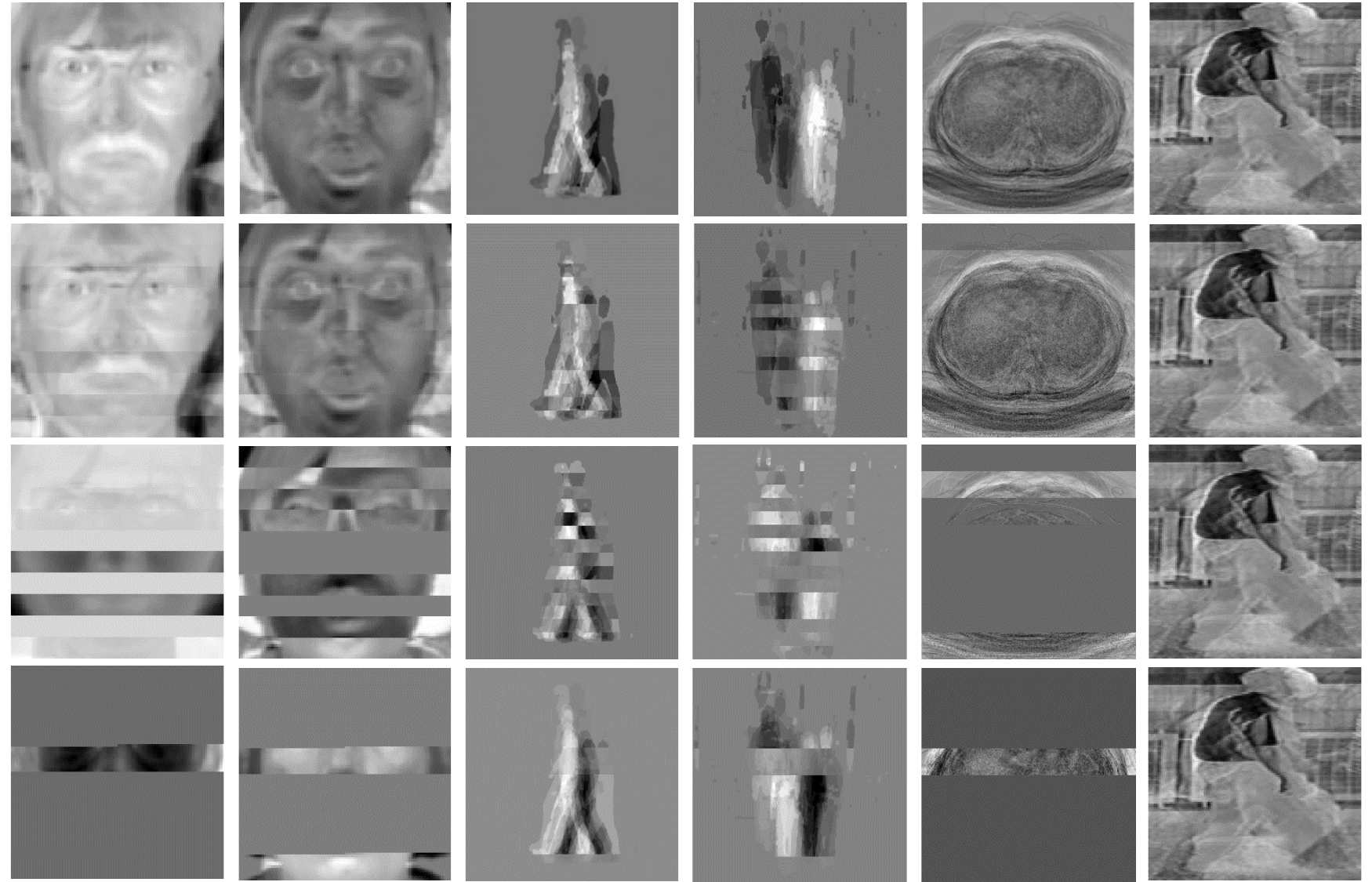}
}
%\caption{The results of comparative experiment on image datasets: YaleFace (center-light and surprised), CasiaGait (sequence 1 and 10), DeepLesion and CUHK03 (from left to right) with algorithms: (a) VFedPCA with central coordination server on the isolated data, fully decentralized VFedPCA on the isolated data, ring topology decentralized VFedPCA on the isolated data, start topology decentralized VFedPCA on the isolated data, from top to bottom; (b) VFedAKPCA with central coordination server on the isolated data, fully decentralized VFedAKPCA on the isolated data, ring topology decentralized VFedAKPCA on the isolated data, start topology decentralized VFedAKPCA on the isolated data, from top to bottom.}
\caption{The results of VFedPCA with the central coordination server, fully decentralized, ring decentralized, and start decentralized, from top to bottom respectively, on the isolated data from each clients.}
\label{fig:example8}
\end{figure}

\item\textbf{The evaluation of two architecture:}
We evaluate the VFedPCA and VFedAKPCA used in the sever-clients architecture and decentralized architecture on different case studies separately. In particular, Figure \ref{fig:example8-1} shows that three different types of decentralized topology. Figure \ref{fig:example8}~(a) shows that the two model are achieve almost the same final result.

\end{itemize}
\color{black}
\section{Conclusion}\label{5}
In this paper, we have proposed VFedPCA and VFedAKPCA algorithm for the linear and nonlinear relationship between various data, which can obtain a collaborative model that improves over the local models learned separately by each client. In addition, we also propose two strategies of the local power iteration warm start method and the weight scaling method to further improve the performance and accuracy of the model. In terms of the communication topology, we have considered both the Server-Client communication topology and the fully decentralized communication topology, where the former is simpler to use and relatively common in FL setting, while the latter provides more flexibility in the communication topology and eliminates the need of the semi-trusted central server. Through extensive comparative studies on various tasks, we have verified that the collaborative model achieves comparative accuracy with the centralized model as if the dataset were un-splitted.

% \section{References}
% \clearpage
\bibliographystyle{IEEEtran}
\bibliography{./ref}}

% \newpage

\appendix
\section{Proof of Theorem 1}\label{6}
Let $[\bm{x}_{1},\bm{x}_{2}\cdots\bm{x}_{n}] \in \mathbb{R}^{\mathnormal{n}}$ be n data points. Then, $\mathbf{A}_{i}\in\mathbb{R}^{\mathnormal{n}\times\mathnormal{n}}$ is a symmetric definite matrix with eigenvalues $\alpha_{1}\geq\alpha_{2}\cdots\geq\alpha_{n}$ and normalized eigenvectors $\mathbf{a}_{1}, \cdots, \mathbf{a}_{n}$, $\Delta=\alpha_{1}-\alpha_{2}$ is the eigen-gap. The local power method estimates the top eigenvector by repeatedly applying the update step $\mathbf{a}_{i}^{l+1} = \frac{\mathbf{A}_{i}\mathbf{a}_{i}^{l}}{\|\mathbf{A}_{i}\mathbf{a}_{i}^{l}\|}$ with an initial vector $\mathbf{{a}_{i}}^0$.
\begin{proof}
We follow the proof of \cite{App}. Define $\theta^{l+1} \in [0, \pi/2]$ by
\begin{equation}
\cos(\theta^{l+1}) =|\mathbf{a}_{1}\mathbf{a}_{i}^{(t+1)}|.
\end{equation}
If $\cos(\theta^{0})\ne0$, then for $\ l \in1,2, \cdots$, we have
\begin{equation}
|\sin(\theta^{l+1})| \le  \tan(\theta^{0})|\frac{\alpha_{2}}{\alpha_{1}}|^{l}
\end{equation}
\begin{equation}
|\alpha^{l}-\alpha_{1}|\le \max_{2\le i\le n}|\alpha_{1}-\alpha_{i}|\tan(\alpha^{0})^2|\frac{\alpha_{2}}{\alpha_{1}}|^{2l}
\end{equation}
During the phase of local power iteration, it follows that $\mathbf{a}_{i}^{l+1}$ is a product of $\mathbf{A}_{i}^{l+1}\mathbf{a}_{i}^{0}$ and so
\begin{equation}
\sin(\theta^{2}) =1- (\mathbf{a}_{1}^T\mathbf{a}_{i}^{(l+1)})^2 =1- (\frac{\mathbf{a}_{1}^T\mathbf{A}_{i}^{l+1}\mathbf{a}_{i}^{0}}{\|\mathbf{A}_{i}^{l+1}\mathbf{a}_{i}^{0}\|} )^2.
\end{equation}
If $\mathbf{a}_{i}^{0}$ has the eigenvector expansion $\mathbf{a}_{i}^{0} = \lambda_1\mathbf{a}_{1}+\cdots+ \lambda_n\mathbf{a}_{n}$, then
\begin{equation}
|\lambda_1|=\cos(\theta^{0})\ne0
\end{equation}
and
\begin{equation}
\mathbf{A}_{i}^{l+1}\mathbf{a}_{i}^{0} = \lambda_{1}\alpha_{1}^{l}\mathbf{a}_{1}+\cdots+\lambda_{n}\alpha_{n}^{l}\mathbf{a}_{n}.
\end{equation}
Therefore, we have
\begin{equation}
\begin{split}
|\sin(\theta)|^{2} &= 1-\frac{\lambda_{1}^{2}\alpha_{1}^{2l}}{\sum_{i=1}^{n}\lambda_{i}^{2}\alpha_{i}^{2l}}\\
&\le \frac{1}{\lambda_{1}^{2}}(\sum_{i=2}^{n}\lambda_{i}^{2})(\frac{\alpha_{2}}{\alpha_{1}})^{2l} \\
&= \tan(\theta^{0})(\frac{\alpha_{2}}{\alpha_{1}})^{2l}.
\end{split}
\end{equation}
Likewise, we have
\begin{equation}
\begin{split}
\alpha^{l}=(\mathbf{a}^{l})^T\mathbf{A}_{i}\mathbf{a}^{l}&=\frac{(\mathbf{a}^{0})^T\mathbf{A}_{i}^{2l+1}\mathbf{a}^{0}}{(\mathbf{a}^{0})^T\mathbf{A}_{i}^{2l}\mathbf{a}^{0}}\\
&=\frac{\sum_{i=1}^{n}\lambda_{i}^{2}\alpha_{i}^{2l+1}}{\sum_{i=1}^{n}\lambda_{i}^{2}\alpha_{i}^{2l}}
\end{split}
\end{equation}
As a result, we have
\begin{equation}
\begin{split}
|\alpha^{l}-\alpha_{1}|&=|\frac{\sum_{i=2}^{n}\lambda_{i}^{2}\alpha_{i}^{2l}(\alpha_{n}-\alpha_{1})}{\sum_{i=1}^{n}\lambda_{i}^{2}\alpha_{i}^{2l}}|\\
&\le \max_{2\le i \le n} |\alpha_{1}-\alpha_{n}|\cdot \tan(\theta^{0})(\frac{\alpha_{2}}{\alpha_{1}})^{2l}
\end{split}
\end{equation}
Thus, after $O(\frac{1}{\Delta}\log{\frac{n}{\epsilon}})$ iterations, the local power method can achieve $\epsilon$-accuracy.

% The i-client per-iteration computational cost is $O(n^2)$, after a total of \(\textit{L}\) iterations, the local power method can converge, so it cost $O(n^2*L)$ total local power iteration computational complexity. During the federated process, the central coordination server aggregates \(\mathbf{a}_{i}^{\textit{l}}\), for all \(i \in1,2, \cdots \textit{p}\), to achieve \(\mathbf{a}^{\textit{l}}\), so it cost $O(p*T)$ total federated communication computational complexity.
\end{proof}
\end{document}